\documentclass[10pt,a4paper]{article}
\usepackage{aria}

\usepackage[utf8]{inputenc}
\usepackage[T1]{fontenc}
\usepackage{hyperref}
\usepackage{url}
\usepackage{microtype}
\usepackage{xcolor}
\usepackage{graphicx}
\usepackage{subcaption}
\usepackage{booktabs}
\usepackage{array}
\usepackage{tabularx}
\usepackage{needspace}
\usepackage{algorithm}
\usepackage{amsmath}
\usepackage{amssymb}
\usepackage{amsfonts}
\usepackage{mathtools}
\usepackage{amsthm}
\usepackage{soul}
\usepackage{enumitem}
\usepackage{multirow}
\usepackage{makecell}
\usepackage{tablefootnote}
\usepackage[para,online,flushleft]{threeparttable}

\setlist{label=\textbullet}

\theoremstyle{plain}

\theoremstyle{definition}

\theoremstyle{remark}

\definecolor{myred}{HTML}{F54254}
\definecolor{myblue}{HTML}{598BE7}

\title{DeFlow: Decoupling Behavior-Prior Modeling and Value Maximization for Offline Policy Extraction}

\author{Zhancun Mu \and Chi Zhang}
\shorttitle{DeFlow}
\affiliation{School of Intelligence Science and Technology\\Peking University}
\correspondence{\texttt{muzhancun@stu.pku.edu.cn}\quad\texttt{chizhang.cz@pku.edu.cn}}
\paperwebsite{https://muzhancun.github.io/publications/deflow/}{muzhancun.github.io/publications/deflow/}

\begin{document}

\maketitle
\fancypagestyle{ariapublished}{%
  \fancyhf{}
  \renewcommand{\headrulewidth}{0pt}
  \fancyfoot[C]{\parbox[b]{\textwidth}{%
    {\color{AriaRule}\hrule height 0.4pt}\vspace{6pt}%
    \raggedright\footnotesize\color{AriaMuted}%
    Published at the 40th Conference on Neural Information Processing Systems (NeurIPS 2026).}}
}
\thispagestyle{ariapublished}

\begin{abstract}
We present DeFlow, a decoupled offline RL framework for extracting high-value actions from a learned multi-step flow prior.
Directly optimizing iterative generative policies typically requires backpropagation through ODE solvers, while shortcut policies trade off expressivity, inference cost, and policy-improvement objectives.
DeFlow keeps the flow model as a behavior prior and trains a lightweight action-conditioned residual module for value improvement under an adaptive trust-region penalty.
This design bypasses solver differentiation and regularizes the refined policy toward the learned behavior prior.
Empirically, DeFlow is competitive with recent generative offline RL baselines, with the clearest gains on selected multimodal manipulation and offline-to-online settings.

% \footnotesize
% \url{http://muzhancun.github.io/preprints/deflow}
\end{abstract}

\begin{figure}[H]
    \centering
    \includegraphics[width=0.68\linewidth]{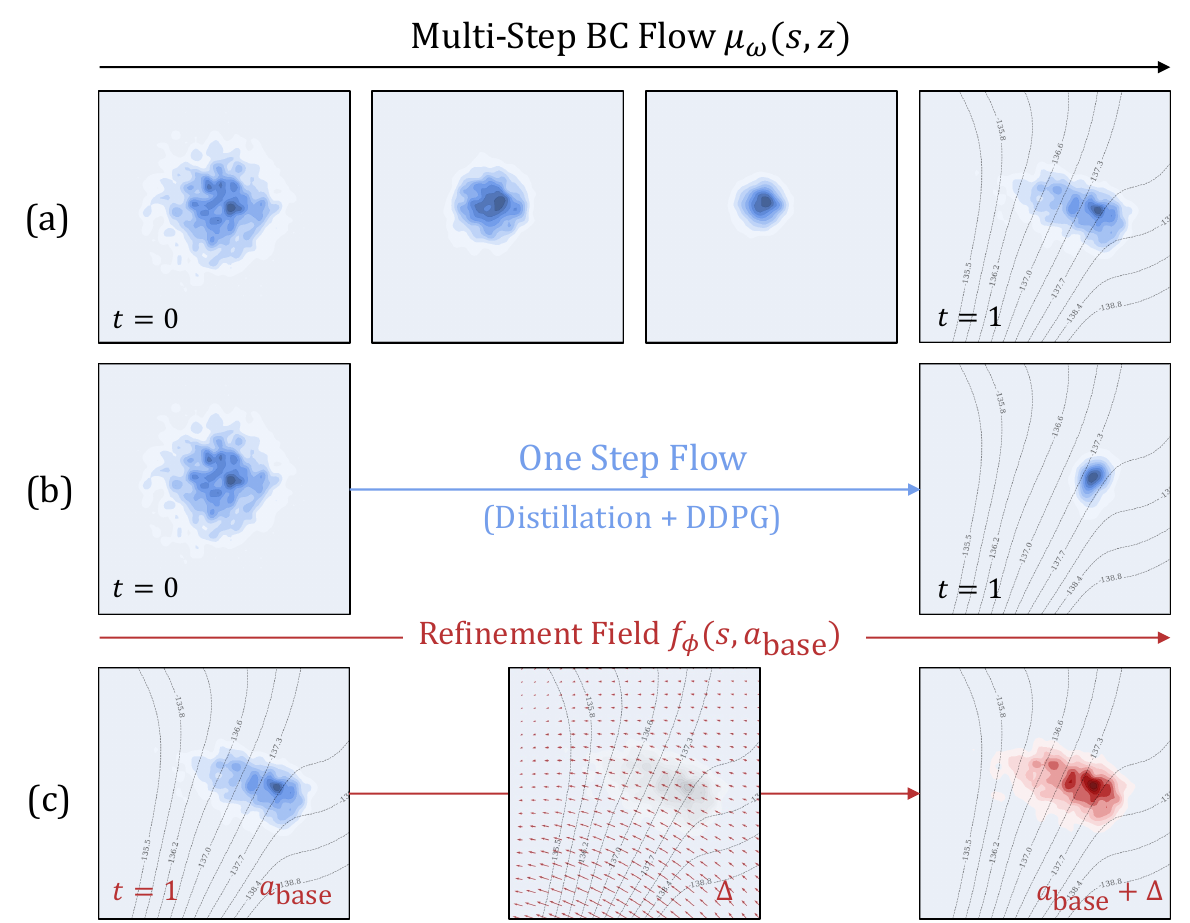}
    \caption{
    \textbf{DeFlow Framework.}
    \textbf{(a)} \textbf{Iterative Flow Prior:} A multi-step flow policy models multimodal behavior.
    \textbf{(b)} \textbf{Shortcut Extraction:} One-step distillation enables direct Q-optimization but may compress the generated action distribution.
    \textbf{(c)} \textbf{DeFlow:} We keep the iterative prior (Blue) and train a lightweight residual module (Orange) to perform local value improvement while penalizing large deviations from the prior.
    }
    \label{fig:deflow}
\end{figure}
\mainmatter

\section{Introduction}

Offline reinforcement learning (RL) must improve a policy while avoiding actions that are weakly supported by the static dataset~\citep{levine2020offline}.
This becomes harder on large multimodal benchmarks such as OGBench~\citep{ogbench_park2025}, where standard unimodal actors often cannot represent the diverse behavior distribution.
Diffusion and flow-matching policies~\citep{flow_lipman2023,diffusion_sohl2015} are attractive in this regime because their iterative sampling procedures can model complex action distributions.

However, extracting a high-value policy from these generative models presents a stark \textit{expressivity-optimizability dilemma}.
Directly differentiating the Q-objective through iterative generation requires solver backpropagation, which can be costly; alternative objectives such as Q-score matching avoid this full differentiation~\citep{psenka2023learning}.
Shortcut methods such as Flow Q-Learning (FQL)~\citep{fql_park2025} and SORL~\citep{espinosa2025scaling} reduce this burden through one-step or shortcut parameterizations, sometimes together with inference-time scaling.
These methods are strong baselines, but they also motivate a complementary question: can we keep a multi-step behavior prior intact and perform value improvement through a small local extraction module?

In this work, we study this decoupled extraction route.
We introduce {\color{myred}DeFlow}, a framework that separates behavior modeling from value-guided policy refinement.

Specifically, DeFlow comprises two distinct components: 
(1) a \textbf{Flow Matching Policy} optimized via supervised behavior modeling; and (2) a lightweight \textbf{Action Refinement Module} that maximizes Q-values via deterministic policy gradients (DDPG~\citep{ddpg_lillicrap2016}).
Instead of updating the flow model through the ODE solver, the refinement module learns an action-conditioned residual shift on the generated action (\Cref{fig:deflow}\textbf{c}).

A critical challenge in this decoupled setting is discouraging large residual moves that exploit unreliable critic values far from the learned prior.
We address this with an adaptive Lagrangian penalty on the residual norm.
The constraint should be interpreted as a trust-region surrogate around the learned flow prior, not as an exact guarantee of support preservation.

Some recent works~\citep{xiao2025selfimproving,dong2025expo} also decouple large expressive policies and small RL heads for policy improvement.
DeFlow instantiates this pattern for offline policy extraction with a flow-matching prior and an instance-level residual conditioned on the sampled action.

Our contribution is the specific combination of a multi-step behavior prior, proposal-conditioned residual extraction, and adaptive residual regularization for offline RL.
Flow matching, residual learning, dual adaptation, and BPTT avoidance are established ingredients rather than separate novelty claims.
We study how this combination behaves relative to shortcut extraction, with benchmark comparisons and diagnostics of residual attribution, capacity, and budget sensitivity.

The results support a scoped claim: DeFlow is competitive with recent generative offline RL methods, while its strongest gains appear in selected multimodal manipulation tasks and in offline-to-online settings where the same residual extraction protocol is reused.

\section{Preliminaries}

We consider the standard Markov Decision Process (MDP)~\citep{mdp_puterman2014} defined by the tuple $(\mathcal{S}, \mathcal{A}, r, p, \gamma)$, where $\mathcal{S}$ and $\mathcal{A}$ denote the state and action spaces, $r(s,a)$ the reward function, $p(s'|s,a)$ the transition dynamics, and $\gamma \in [0,1)$ the discount factor.

\subsection{Offline RL as Constrained Optimization}

In offline RL, we are given a static dataset $\mathcal{D} = \{(s, a, r, s')\}$ collected by a behavior policy $\pi_\beta$. The goal is to learn a policy $\pi_\theta$ that maximizes the expected cumulative return. However, due to the distributional shift, simply maximizing values on out-of-distribution (OOD) actions leads to extrapolation errors. Following \citet{levine2020offline}, offline RL is often cast as a constrained optimization problem:
\begin{equation}
\begin{aligned}
    \max_{\pi_\theta}\, &\mathbb{E}_{s \sim \mathcal{D}, a \sim \pi_\theta(\cdot|s)} [Q(s, a)] \\ \text{s.t.} \quad &\mathbb{D}(\pi_\theta(\cdot|s) || \pi_\beta(\cdot|s)) \le \epsilon,
\end{aligned}
\label{eq:constrained_obj}
\end{equation}
where $Q(s,a)$ is the state-action value function and $\mathbb{D}$ is a divergence measure (e.g., KL divergence or Wasserstein distance) used to discourage unsupported policy shifts.

\paragraph{Critic Learning.}
To evaluate the policy, we learn a critic $Q_\xi$ by minimizing the standard Bellman error~\citep{dqn_mnih2013}. The objective is:

\begin{equation}
\begin{aligned}
\mathcal{L}_Q(\xi)
&=
\mathbb{E}_{(s,a,r,s') \sim \mathcal{D}}
\Bigl[
    \bigl(
    Q_\xi(s,a)-r-\gamma \hat Q_{\bar\xi}(s',a')
    \bigr)^2
\Bigr].
\end{aligned}
\end{equation}
where $a' \sim \pi_\theta(\cdot|s')$ and $\hat{Q}_{\bar{\xi}}$ utilizes a target network for stability. Critic parameters $\xi$, prior parameters $\psi$, and residual parameters $\phi$ are distinct.

\paragraph{The Actor's Dilemma.} Standard methods typically convert the constrained problem in Eq. (\ref{eq:constrained_obj}) into an unconstrained one via a Lagrangian relaxation with a fixed penalty coefficient $\alpha$:
\begin{equation}
\begin{aligned}
\mathcal{L}_\pi(\theta)
&=
\mathbb{E}_{s \sim \mathcal{D}}
\Bigl[
    -\mathbb{E}_{a \sim \pi_\theta}[Q_\xi(s,a)]
    + \alpha\,\mathcal{L}_{\mathrm{BC}}(\pi_\theta,\mathcal{D})
\Bigr].
\end{aligned}
\label{eq:standard_actor}
\end{equation}
Gaussian policies allow efficient reparameterized actor updates but have limited ability to represent multimodal behavior. Iterative generative policies offer greater expressivity, while direct differentiation through their sampling chains increases the optimization cost. Alternative training objectives can avoid this differentiation. A fixed regularization coefficient also introduces a task-dependent trade-off between value improvement and prior proximity~\citep{fql_park2025,agrawalla2025floq}.

\subsection{Conditional Flow Matching}

To address the modeling challenge, we leverage Conditional Flow Matching (CFM)~\citep{flow_lipman2023}, which offers a simulation-free alternative to diffusion training. CFM learns a time-dependent vector field $v_\psi(t, x): [0,1] \times \mathbb{R}^d \to \mathbb{R}^d$ that pushes a simple prior distribution $p_0$ (e.g., $\mathcal{N}(0, I)$) to the data distribution $p_1$ (e.g., $\pi_\beta(a|s)$).

The generative process is defined by an Ordinary Differential Equation (ODE):
\begin{equation}
\frac{\mathrm{d}}{\mathrm{d}t}\Psi_t(x) = v_\psi(t, \Psi_t(x)), \quad \Psi_0(x) = x \sim p_0,
\end{equation}
where $\Psi_t$ is the flow map. We adopt the straight conditional path $x_t = (1-t)x_0 + t x_1$~\citep{flow_lipman2024}. Noise and data endpoints are sampled independently; this does not assert an optimal coupling between the endpoint distributions. The vector field is trained via the regression objective:
\begin{equation}
\begin{aligned}
\mathcal{L}_{\mathrm{FM}}(\psi)
&=
\mathbb{E}_{t \sim \mathcal{U}[0,1],\,x_0 \sim p_0,\,x_1 \sim \mathcal{D}}
\bigl\|
v_\psi(t,x_t)-(x_1-x_0)
\bigr\|_2^2.
\end{aligned}
\label{eq:flow}
\end{equation}

\paragraph{Flow as a Policy.} In our context, the flow model serves as a state-conditional generator. We denote the generated action as a deterministic function of the initial noise $z$ and state $s$:
\begin{equation}
a_{\text{flow}} = \text{ODESolve}(v_\psi, s, z), \quad z \sim \mathcal{N}(0, I).
\end{equation}
In our framework, this flow model acts as the learned behavior prior, decoupling distribution modeling from the policy improvement step described next.

\section{DeFlow: Disentangling Fidelity and Optimality}

\begin{algorithm}[t!]
\caption{DeFlow}
\label{alg:deflow}
\small
\begin{algorithmic}[1]
\Function{$\mu_\psi(s,z)$}{}
    \For{$m=0,\dots,M-1$}
        \State $z \gets z + v_\psi(m/M,s,z)/M$
    \EndFor
    \State \Return $\operatorname{clip}(z,-1,1)$
\EndFunction
\Function{Act}{$s$}
    \State $z\sim\mathcal{N}(0,I_d)$,\quad $a_b\gets\mu_\psi(s,z)$
    \State \Return $\operatorname{clip}(a_b+f_\phi(s,a_b),-1,1)$
\EndFunction
\While{not converged}
    \State Sample minibatch $\{(s,a,r,s')\}\sim\mathcal{D}$
    \State Update $Q_\xi$ with Bellman targets using $a'=\Call{Act}{s'}$
    \State Update $v_\psi$ on $\|v_\psi(t,s,x^t)-(x^1-x^0)\|_2^2$
    \State $a_b\gets\operatorname{sg}[\mu_\psi(s,z)]$, \quad $\Delta\gets f_\phi(s,a_b)$
    \State Update $f_\phi$ on $-Q_\xi^{\mathrm{norm}}(s,\operatorname{clip}(a_b+\Delta,-1,1))+\operatorname{sg}[\alpha]\|\Delta\|_2^2/d$
    \State Update $\eta=\log\alpha$ on $-\eta\,\operatorname{sg}(\|\Delta\|_2^2/d-\delta)$
\EndWhile
\Return $\mu_\psi,f_\phi$
\end{algorithmic}
\end{algorithm}

DeFlow disentangles behavior modeling from value-guided extraction.
Instead of training one generative policy with both behavior-cloning and Q-maximization gradients, we keep the multi-step flow as a proposal distribution and learn a small residual extractor on top of its samples.
\Cref{alg:deflow} provides an overview of our method.

We formulate the policy as a composite function, where the final action $a$ is derived from a \textbf{Base Flow} $\mu_\psi$ adjusted by a \textbf{Refinement Module} $f_\phi$:
\begin{equation}
a = \underbrace{\text{sg}[\mu_\psi(s, z)]}_{\text{Fixed Proposal}} + f_\phi(s, \underbrace{\text{sg}[\mu_\psi(s, z)]}_{\text{Context}}), \quad z \sim \mathcal{N}(0, I)
\label{eq:deflow_formulation}
\end{equation}
Here, $\text{sg}[\cdot]$ denotes the stop-gradient operator, and the displayed action is before clipping to the action bounds. The flow $\mu_\psi$ provides a behavior-prior proposal, while $f_\phi$, implemented as an MLP, learns a residual shift toward higher critic values.

\textbf{The role of Stop-Gradient:} The stop-gradient operator cuts the computational graph between the refinement module and the flow model. During the update of $f_\phi$, the base flow $\mu_\psi$ is treated as a fixed stochastic generator rather than a differentiable layer. This bypasses backpropagation through the ODE solver and keeps the two optimization loops separated.

\subsection{Revisiting Existing Coupling Paradigms}

To better contextualize DeFlow, we briefly analyze the limitations inherent in two prevalent strategies for integrating flow models with Q-learning.

\paragraph{Paradigm I: End-to-End Policy Steering.} 
Ideally, one might update a flow model $\pi_\psi$ by backpropagating the Q-gradient through the ODE integration chain:
\begin{equation}
\nabla_\psi \mathcal{L}_{\text{RL}} = \mathbb{E}_{s, z} \left[ \frac{\partial Q(s, a)}{\partial a} \cdot \frac{\partial \text{ODESolve}(v_\psi, s, z)}{\partial \psi} \right].
\end{equation}
While conceptually direct, this ``steering'' approach requires BPTT or the Adjoint Method across numerous integration steps. This is memory-intensive and can suffer from unstable or weak gradients through the solver.

\paragraph{Paradigm II: Single-step Shortcut Distillation.}
Alternatively, methods like FQL~\citep{fql_park2025} attempt to bypass the ODE complexity by distilling the iterative flow into a single-step policy $\pi_\theta(s, z)$. The actor is typically trained with a composite objective:
\begin{align}
    \mathcal{L}_\pi(\omega) &=
    \underbrace{\mathbb{E}_{s, a^\pi}[-Q_\xi(s, a^\pi)]}_{\texttt{Q maximization}}
    + \underbrace{\alpha \mathcal{L}_\mathrm{Distill}(\omega)}_{\texttt{BC regularization}}. \label{eq:fql_actor}
\end{align}
While efficient at inference, this paradigm can compress the complex, multimodal distribution represented by the original flow. Furthermore, the scalar $\alpha$ introduces a sensitive hyperparameter trade-off: a large $\alpha$ limits improvement, while a small $\alpha$ allows larger deviations from the behavior prior.

\subsection{The DeFlow Approach}

DeFlow addresses these challenges by maintaining the \textit{Multi-step Flow} as a behavior prior while employing a \textit{Residual Refinement} for optimization.

The Base Flow $\mu_\psi$ is updated solely via the Flow Matching objective $\mathcal{L}_{\text{FM}}(\psi)$, so Q-learning gradients do not directly alter the prior.

Inspired by trust-region methods, we formulate refinement using a fixed, task-level budget $\delta$ on the mean squared residual per action coordinate. Let $d$ be the action dimension:
\begin{equation}
\max_{f_\phi} \mathbb{E}_{s, z} [Q(s, a_{\text{flow}} + \Delta a)] \quad \text{s.t.} \quad \mathbb{E}_{s, z} [\|\Delta a\|_2^2/d] \le \delta,
\label{eq:op}
\end{equation}
where $\Delta a = f_\phi(s, a_{\text{flow}})$ and $a_{\text{flow}} = \text{sg}[\mu_\psi(s, z)]$. Here $s$ is sampled from replay. The division by $d$ matches the implementation's coordinate-averaged squared error; $\delta$ is a squared-displacement budget, not a per-action radius. We apply action clipping when evaluating the critic and executing the policy.
From an optimal-transport view~\citep{ot_villani2009,computational_ot_peyre2019}, the residual map $T_s(a_b)=a_b+f_\phi(s,a_b)$ is a Monge-style transport from the learned flow prior to the refined policy. Under the coupling induced by sampling $a_b\sim\mu_\psi(\cdot|s)$ and mapping it through $T_s$, $\mathbb{E}\|\Delta a\|_2^2$ upper-bounds the squared $W_2$ distance to the learned prior; we state this formally in \cref{app:ot}.

\paragraph{Relation to residual policies and SORL.}
Classical residual policy learning~\citep{silver2019residual} usually adds a state-dependent correction to a base controller.
DeFlow instead uses an instance-level residual $f_\phi(s,a_b)$, allowing different sampled actions from the same state to receive different local corrections.
SORL~\citep{espinosa2025scaling} is an important step-size-conditioned shortcut-flow baseline: its actor is trained with flow matching, shortcut-consistency, and Q-maximization losses, and its evaluation can sweep shortcut step counts with optional critic-ranked candidates.
We include it in the main benchmark, while DeFlow focuses on local residual extraction around a prior held fixed during each actor update. The prior continues to receive its separate behavior-modeling updates.
Action conditioning is not a mode classifier: different proposals can receive different corrections, which may be mode-specific when modes are separated in action space.

To solve \cref{eq:op}, we employ the method of Lagrange multipliers. We first apply \textbf{Q-Normalization} to the RL objective to mitigate the scale variance of $Q$-values across different environments. We then introduce a learnable Lagrangian multiplier $\alpha$ (where $\alpha > 0$). The loss function for the refinement module $f_\phi$ is defined as:
\begin{equation}
\begin{aligned}
\mathcal{L}_{\mathrm{Refine}}(\phi)
&=
\mathbb{E}_{s,z}
\Biggl[
    -\frac{
        Q(s,\operatorname{clip}(a_{\mathrm{flow}}+\Delta a,-1,1))
    }{
        \operatorname{sg}(\mathbb{E}_{\mathrm{batch}}|Q|)
    }
    + \operatorname{sg}[\alpha]\|\Delta a\|_2^2/d
\Biggr].
\end{aligned}
\end{equation}

We adapt the positive penalty $\alpha=\exp(\eta)$ using a log-space dual update. In the implementation, the loss minimized with respect to $\eta$ is:
\begin{equation}
\mathcal{L}_{\mathrm{dual}}(\eta) = \eta\left(\delta-\operatorname{sg}\left[\mathbb{E}_{s,z}\|\Delta a\|_2^2/d\right]\right).
\label{eq:alpha_loss}
\end{equation}
When the empirical residual exceeds $\delta$, gradient descent increases $\eta$ and hence $\alpha$; otherwise it decreases the penalty. This resembles SAC's log-temperature adaptation~\citep{sac}, but controls residual movement rather than entropy. It is a surrogate dual update, not a claim of solving the nonconvex constrained problem exactly. Adaptive $\alpha$ reduces fixed-coefficient tuning; $\delta$ still requires selection. Minibatch regularization neither enforces a hard pointwise bound nor guarantees true behavior-support membership.

\section{Experiments}
\label{sec:exp}

We evaluate DeFlow on \textbf{$73$} offline RL tasks and \textbf{$15$} offline-to-online (O2O) tasks.
The goal is to test whether local residual extraction around a multi-step flow prior is competitive with recent generative offline RL methods, and where it helps or underperforms.

\begin{table*}[t!]
\caption{
\textbf{Offline RL results on OGBench and D4RL.}
We report DeFlow against Gaussian-policy and flow-based/generative policy-extraction baselines.
SORL results are included where aligned OGBench numbers are available; missing entries are marked by ``-'' and are not used when identifying row-wise best scores.
Results are averaged over 8 seeds for state tasks and 4 seeds for visual tasks when applicable.
Bold entries are at least $0.95\times$ the best reported score in their row.
See \Cref{table:offline_full} for the full task-level version of the non-SORL baselines.
Counts in parentheses indicate the number of tasks.
}
\label{table:offline}
\centering
\fontsize{7.5}{9}\selectfont
\setlength{\tabcolsep}{2pt}
\begin{threeparttable}
\begin{tabular*}{\textwidth}{@{\extracolsep{\fill}}lccccccccc@{}}
\toprule
\multicolumn{1}{c}{} & \multicolumn{3}{c}{\texttt{Gaussian Policies}} & \multicolumn{6}{c}{\texttt{Flow / Generative Extraction}} \\
\cmidrule(lr){2-4} \cmidrule(lr){5-10}
\texttt{Task Category} & \texttt{BC} & \texttt{IQL} & \texttt{ReBRAC} & \texttt{FAWAC} & \texttt{FBRAC} & \texttt{IFQL} & \texttt{SORL} & \texttt{\color{myblue}FQL} & \texttt{\color{myred}DeFlow}\\
\midrule
\texttt{antmaze-large (5)} & $11$ {\tiny $\pm 1$} & $53$ {\tiny $\pm 3$} & $81$ {\tiny $\pm 5$} & $6$ {\tiny $\pm 1$} & $60$ {\tiny $\pm 6$} & $28$ {\tiny $\pm 5$} & $\mathbf{89}$ {\tiny $\pm 2$} & $79$ {\tiny $\pm 3$} & $81$ {\tiny $\pm 3$} \\
\texttt{antmaze-giant (5)} & $0$ {\tiny $\pm 0$} & $4$ {\tiny $\pm 1$} & $\mathbf{26}$ {\tiny $\pm 8$} & $0$ {\tiny $\pm 0$} & $4$ {\tiny $\pm 4$} & $3$ {\tiny $\pm 2$} & $9$ {\tiny $\pm 6$} & $9$ {\tiny $\pm 6$} & $12$ {\tiny $\pm 5$}\\
\texttt{humanoidmaze-medium (5)} & $2$ {\tiny $\pm 1$} & $33$ {\tiny $\pm 2$} & $22$ {\tiny $\pm 8$} & $19$ {\tiny $\pm 1$} & $38$ {\tiny $\pm 5$} & $60$ {\tiny $\pm 14$} & $\mathbf{64}$ {\tiny $\pm 4$} & $58$ {\tiny $\pm 5$} & $48$ {\tiny $\pm 4$} \\
\texttt{humanoidmaze-large (5)} & $1$ {\tiny $\pm 0$} & $2$ {\tiny $\pm 1$} & $2$ {\tiny $\pm 1$} & $0$ {\tiny $\pm 0$} & $2$ {\tiny $\pm 0$} & $\mathbf{11}$ {\tiny $\pm 2$} & $5$ {\tiny $\pm 2$} & $4$ {\tiny $\pm 2$} & $5$ {\tiny $\pm 2$}\\
\texttt{antsoccer-arena (5)} & $1$ {\tiny $\pm 0$} & $8$ {\tiny $\pm 2$} & $0$ {\tiny $\pm 0$} & $12$ {\tiny $\pm 0$} & $16$ {\tiny $\pm 1$} & $33$ {\tiny $\pm 6$} & $\mathbf{69}$ {\tiny $\pm 2$} & $60$ {\tiny $\pm 2$} & $\mathbf{67}$ {\tiny $\pm 3$}\\
\texttt{cube-single (5)} & $5$ {\tiny $\pm 1$} & $83$ {\tiny $\pm 3$} & $91$ {\tiny $\pm 2$} & $81$ {\tiny $\pm 4$} & $79$ {\tiny $\pm 7$} & $79$ {\tiny $\pm 2$} & $\mathbf{97}$ {\tiny $\pm 1$} & $\mathbf{96}$ {\tiny $\pm 1$} & $\mathbf{96}$ {\tiny $\pm 2$}\\
\texttt{cube-double (5)} & $2$ {\tiny $\pm 1$} & $7$ {\tiny $\pm 1$} & $12$ {\tiny $\pm 1$} & $5$ {\tiny $\pm 2$} & $15$ {\tiny $\pm 3$} & $14$ {\tiny $\pm 3$} & $25$ {\tiny $\pm 3$} & $29$ {\tiny $\pm 2$} & $\mathbf{40}$ {\tiny $\pm 5$}\\
\texttt{scene (5)} & $5$ {\tiny $\pm 1$} & $28$ {\tiny $\pm 1$} & $41$ {\tiny $\pm 3$} & $30$ {\tiny $\pm 3$} & $45$ {\tiny $\pm 5$} & $30$ {\tiny $\pm 3$} & $\mathbf{57}$ {\tiny $\pm 2$} & $\mathbf{56}$ {\tiny $\pm 2$} & $51$ {\tiny $\pm 3$}\\
\texttt{puzzle-3x3 (5)} & $2$ {\tiny $\pm 0$} & $9$ {\tiny $\pm 1$} & $21$ {\tiny $\pm 1$} & $6$ {\tiny $\pm 2$} & $14$ {\tiny $\pm 4$} & $19$ {\tiny $\pm 1$} & $-$ & $30$ {\tiny $\pm 1$} & $\mathbf{43}$ {\tiny $\pm 4$}\\
\texttt{puzzle-4x4 (5)} & $0$ {\tiny $\pm 0$} & $7$ {\tiny $\pm 1$} & $14$ {\tiny $\pm 1$} & $1$ {\tiny $\pm 0$} & $13$ {\tiny $\pm 1$} & $\mathbf{25}$ {\tiny $\pm 5$} & $-$ & $17$ {\tiny $\pm 2$} & $11$ {\tiny $\pm 2$} \\
\texttt{D4RL antmaze (6)} & $17$ & $57$ & $78$ & $44$ {\tiny $\pm 3$} & $64$ {\tiny $\pm 7$} & $65$ {\tiny $\pm 7$} & $-$ & $\mathbf{84}$ {\tiny $\pm 3$} & $\mathbf{83}$ {\tiny $\pm 2$}\\
\texttt{D4RL adroit (12)} & $48$ & $53$ & $\mathbf{59}$ & $48$ {\tiny $\pm 1$} & $50$ {\tiny $\pm 2$} & $52$ {\tiny $\pm 1$} & $-$ & $52$ {\tiny $\pm 1$} &$52$ {\tiny $\pm 1$}\\
\texttt{Visual manipulation (5)} & - & $42$ {\tiny $\pm 4$} & $60$ {\tiny $\pm 2$} & - & $22$ {\tiny $\pm 2$} & $50$ {\tiny $\pm 5$} & $-$ & $65$ {\tiny $\pm 2$} & $\mathbf{69}$ {\tiny $\pm 2$}\\
\bottomrule
\end{tabular*}
\begin{tablenotes}
\item[1] Pixel tasks selectively benchmark methods that are strong on state-based OGBench tasks.
\item[2] SORL numbers are taken from the aligned OGBench comparison used in DROL; SORL uses step-size-conditioned shortcut flows and inference-time shortcut-step scaling.
\end{tablenotes}
\end{threeparttable}
\end{table*}

\subsection{Experimental Setup}

\paragraph{Benchmarks.} We use OGBench~\citep{ogbench_park2025} as the primary benchmark, including 50 state-based reward-conditioned \texttt{single-task} tasks and 5 visual manipulation tasks. We also evaluate on 18 D4RL tasks~\citep{d4rl_fu2020}.

\paragraph{Baselines.} We compare against Gaussian-policy baselines (BC, IQL~\citep{iql_kostrikov2022}, ReBRAC~\citep{rebrac_tarasov2023}) and flow-based extraction baselines (FAWAC, FBRAC, IFQL, FQL~\citep{fql_park2025}). We also include SORL~\citep{espinosa2025scaling} wherever aligned OGBench results are available. SORL is not an apples-to-apples matched-inference baseline because its shortcut-flow implementation conditions on step size and evaluates multiple shortcut step counts, with optional best-of-$N$ critic selection; nevertheless, it is an important recent reference point for efficient generative offline RL.

\paragraph{Offline-to-online Experiments.}
For O2O experiments, we further compare against \textbf{Cal-QL}~\citep{calql_nakamoto2023} and \textbf{RLPD}~\citep{rlpd_ball2023}, which are specifically engineered for online adaptation.
We adapt DeFlow for this setting by simply augmenting the offline dataset $\mathcal{D}$ with an online replay buffer during training.

\paragraph{Evaluation Protocol.} We follow the OGBench protocol~\citep{ogbench_park2025}, reporting performance after a fixed number of gradient steps. Results are averaged over 8 seeds for state-based tasks and 4 seeds for pixel-based tasks. We \textbf{bold} scores that are within 95\% of the best reported performance in each row.
The main results use a fixed four-layer residual; we do not select residual depth using the additional diagnostics. Their four-seed protocols and final-checkpoint evaluation budgets are reported separately in \cref{sec:capacity}.

\begin{table*}[t!]
\caption{
\textbf{Offline-to-online RL results.}
The results are averaged over $8$ seeds unless otherwise mentioned.
Entries show offline $\to$ online performance; OGBench rows use the default single-task environment.
}
\label{table:o2o_full}
\centering
\setlength{\tabcolsep}{2.5pt}
\resizebox{\textwidth}{!}{%
\begin{tabular}{lccccccc}

\toprule
\texttt{Task} & \texttt{IQL} & \texttt{ReBRAC} & \texttt{Cal-QL} & \texttt{RLPD} & \texttt{IFQL} & \texttt{\color{myblue}FQL} & \texttt{\color{myred}{DeFlow}}
\\
\midrule

\texttt{humanoidmaze-medium}
& $21$ {\tiny $\pm 13$} $\to$ $16$ {\tiny $\pm 8$} & $16$ {\tiny $\pm 20$} $\to$ $1$ {\tiny $\pm 1$} & $0$ {\tiny $\pm 0$} $\to$ $0$ {\tiny $\pm 0$} & $0$ {\tiny $\pm 0$} $\to$ $8$ {\tiny $\pm 10$} & $56$ {\tiny $\pm 35$} $\to$ $\mathbf{82}$ {\tiny $\pm 20$} & $12$ {\tiny $\pm 7$} $\to$ $22$ {\tiny $\pm 12$} & {\color{myred}$13$ {\tiny $\pm 5$} $\to$ $65$ {\tiny $\pm 12$}} \\
\texttt{antsoccer-arena} & $2$ {\tiny $\pm 1$} $\to$ $0$ {\tiny $\pm 0$} & $0$ {\tiny $\pm 0$} $\to$ $0$ {\tiny $\pm 0$} & $0$ {\tiny $\pm 0$} $\to$ $0$ {\tiny $\pm 0$} & $0$ {\tiny $\pm 0$} $\to$ $0$ {\tiny $\pm 0$} & $26$ {\tiny $\pm 15$} $\to$ $39$ {\tiny $\pm 10$} & $28$ {\tiny $\pm 8$} $\to$ $\mathbf{86}$ {\tiny $\pm 5$} & $44$ {\tiny $\pm 8$} $\to$ $\mathbf{86}$ {\tiny $\pm 3$}\\
\texttt{cube-double} & $0$ {\tiny $\pm 1$} $\to$ $0$ {\tiny $\pm 0$} & $6$ {\tiny $\pm 5$} $\to$ $28$ {\tiny $\pm 28$} & $0$ {\tiny $\pm 0$} $\to$ $0$ {\tiny $\pm 0$} & $0$ {\tiny $\pm 0$} $\to$ $0$ {\tiny $\pm 0$} & $12$ {\tiny $\pm 9$} $\to$ $40$ {\tiny $\pm 5$} & $40$ {\tiny $\pm 11$} $\to$ $\mathbf{92}$ {\tiny $\pm 3$} & $48$ {\tiny $\pm 13$} $\to$ $\mathbf{93}$ {\tiny $\pm 4$}\\
\texttt{scene} & $14$ {\tiny $\pm 11$} $\to$ $10$ {\tiny $\pm 9$} & $55$ {\tiny $\pm 10$} $\to$ $\mathbf{100}$ {\tiny $\pm 0$} & $1$ {\tiny $\pm 2$} $\to$ $50$ {\tiny $\pm 53$} & $0$ {\tiny $\pm 0$} $\to$ $\mathbf{100}$ {\tiny $\pm 0$} & $0$ {\tiny $\pm 1$} $\to$ $60$ {\tiny $\pm 39$} & $82$ {\tiny $\pm 11$} $\to$ $\mathbf{100}$ {\tiny $\pm 1$} & $58$ {\tiny $\pm 14$} $\to$ $\mathbf{100}$ {\tiny $\pm 1$}\\
\texttt{puzzle-4x4} & $5$ {\tiny $\pm 2$} $\to$ $1$ {\tiny $\pm 1$} & $8$ {\tiny $\pm 4$} $\to$ $14$ {\tiny $\pm 35$} & $0$ {\tiny $\pm 0$} $\to$ $0$ {\tiny $\pm 0$} & $0$ {\tiny $\pm 0$} $\to$ $\mathbf{100}$ {\tiny $\pm 1$} & $23$ {\tiny $\pm 6$} $\to$ $19$ {\tiny $\pm 33$} & $8$ {\tiny $\pm 3$} $\to$ $38$ {\tiny $\pm 52$} & {\color{myred}$4$ {\tiny $\pm 4$} $\to$ $\mathbf{100}$ {\tiny $\pm 0$}} \\
\midrule
\texttt{antmaze-umaze-v2} & $77$ $\to$ $\mathbf{96}$ & $98$ $\to$ $75$ & $77$ $\to$ $\mathbf{100}$ & $0$ {\tiny $\pm 0$} $\to$ $\mathbf{98}$ {\tiny $\pm 3$} & $94$ {\tiny $\pm 5$} $\to$ $\mathbf{96}$ {\tiny $\pm 2$} & $97$ {\tiny $\pm 2$} $\to$ $\mathbf{99}$ {\tiny $\pm 1$} & $96$ {\tiny $\pm 3$} $\to$ $\mathbf{99}$ {\tiny $\pm 2$}\\
\texttt{antmaze-umaze-diverse-v2} & $60$ $\to$ $64$ & $74$ $\to$ $\mathbf{98}$ & $32$ $\to$ $\mathbf{98}$ & $0$ {\tiny $\pm 0$} $\to$ $94$ {\tiny $\pm 5$} & $69$ {\tiny $\pm 20$} $\to$ $93$ {\tiny $\pm 5$} & $79$ {\tiny $\pm 16$} $\to$ $\mathbf{100}$ {\tiny $\pm 1$} & $87$ {\tiny $\pm 6$} $\to$ $\mathbf{99}$ {\tiny $\pm 1$} \\
\texttt{antmaze-medium-play-v2} & $72$ $\to$ $90$ & $88$ $\to$ $\mathbf{98}$ & $72$ $\to$ $\mathbf{99}$ & $0$ {\tiny $\pm 0$} $\to$ $\mathbf{98}$ {\tiny $\pm 2$} & $52$ {\tiny $\pm 19$} $\to$ $93$ {\tiny $\pm 2$} & $77$ {\tiny $\pm 7$} $\to$ $\mathbf{97}$ {\tiny $\pm 2$} & $76$ {\tiny $\pm 5$} $\to$ $\mathbf{97}$ {\tiny $\pm 2$}\\
\texttt{antmaze-medium-diverse-v2} & $64$ $\to$ $92$ & $85$ $\to$ $\mathbf{99}$ & $62$ $\to$ $\mathbf{98}$ & $0$ {\tiny $\pm 0$} $\to$ $\mathbf{97}$ {\tiny $\pm 2$} & $44$ {\tiny $\pm 26$} $\to$ $89$ {\tiny $\pm 4$} & $55$ {\tiny $\pm 19$} $\to$ $\mathbf{97}$ {\tiny $\pm 3$} & $61$ {\tiny $\pm 9$} $\to$ $\mathbf{98}$ {\tiny $\pm 2$}\\
\texttt{antmaze-large-play-v2} & $38$ $\to$ $64$ & $68$ $\to$ $32$ & $32$ $\to$ $\mathbf{97}$ & $0$ {\tiny $\pm 0$} $\to$ $\mathbf{93}$ {\tiny $\pm 5$} & $64$ {\tiny $\pm 14$} $\to$ $80$ {\tiny $\pm 5$} & $66$ {\tiny $\pm 40$} $\to$ $84$ {\tiny $\pm 30$} & $76$ {\tiny $\pm 6$} $\to$ $\mathbf{95}$ {\tiny $\pm 4$} \\
\texttt{antmaze-large-diverse-v2} & $27$ $\to$ $64$ & $67$ $\to$ $72$ & $44$ $\to$ $\mathbf{92}$ & $0$ {\tiny $\pm 0$} $\to$ $\mathbf{94}$ {\tiny $\pm 3$} & $69$ {\tiny $\pm 6$} $\to$ $86$ {\tiny $\pm 5$} & $75$ {\tiny $\pm 24$} $\to$ $\mathbf{94}$ {\tiny $\pm 3$} & $76$ {\tiny $\pm 9$} $\to$ $\mathbf{96}$ {\tiny $\pm 2$} \\
\midrule
\texttt{pen-cloned-v1} & $84$ $\to$ $102$ & $74$ $\to$ $138$ & $-3$ $\to$ $-3$ & $3$ {\tiny $\pm 2$} $\to$ $120$ {\tiny $\pm 10$} & $77$ {\tiny $\pm 7$} $\to$ $107$ {\tiny $\pm 10$} & $53$ {\tiny $\pm 14$} $\to$ $\mathbf{149}$ {\tiny $\pm 6$} & $64$ {\tiny $\pm 13$} $\to$ $\mathbf{142}$ {\tiny $\pm 7$}\\
\texttt{door-cloned-v1} & $1$ $\to$ $20$ & $0$ $\to$ $\mathbf{102}$ & $-0$ $\to$ $-0$ & $0$ {\tiny $\pm 0$} $\to$ $\mathbf{102}$ {\tiny $\pm 7$} & $3$ {\tiny $\pm 2$} $\to$ $50$ {\tiny $\pm 15$} & $0$ {\tiny $\pm 0$} $\to$ $\mathbf{102}$ {\tiny $\pm 5$} & $0$ {\tiny $\pm 0$} $\to$ $\mathbf{99}$ {\tiny $\pm 2$}\\
\texttt{hammer-cloned-v1} & $1$ $\to$ $57$ & $7$ $\to$ $\mathbf{125}$ & $0$ $\to$ $0$ & $0$ {\tiny $\pm 0$} $\to$ $\mathbf{128}$ {\tiny $\pm 29$} & $4$ {\tiny $\pm 2$} $\to$ $60$ {\tiny $\pm 14$} & $0$ {\tiny $\pm 0$} $\to$ $\mathbf{127}$ {\tiny $\pm 17$} & $7$ {\tiny $\pm 3$} $\to$ $106$ {\tiny $\pm 10$}\\
\texttt{relocate-cloned-v1} & $0$ $\to$ $0$ & $1$ $\to$ $7$ & $-0$ $\to$ $-0$ & $0$ {\tiny $\pm 0$} $\to$ $2$ {\tiny $\pm 2$} & $-0$ {\tiny $\pm 0$} $\to$ $5$ {\tiny $\pm 3$} & $0$ {\tiny $\pm 1$} $\to$ $\mathbf{62}$ {\tiny $\pm 8$} & $\mathbf{1}$ {\tiny $\pm 0$} $\to$ $35$ {\tiny $\pm 8$} \\

\bottomrule
\end{tabular}
}
\end{table*}

\subsection{Results and Analysis}

\paragraph{Offline results.}
\Cref{table:offline} shows a mixed but informative comparison.
SORL is very strong on navigation and scene-style tasks, including \texttt{antmaze-large}, \texttt{humanoidmaze-medium}, \texttt{antsoccer}, and \texttt{scene}.
DeFlow is most competitive on manipulation tasks where local multimodal precision matters: it improves over FQL and the reported SORL numbers on \texttt{cube-double}, and it remains strong on \texttt{puzzle-3x3} and visual manipulation.
On D4RL, DeFlow is close to FQL on AntMaze and Adroit rather than clearly dominating.
These results support a scoped claim: residual extraction around a multi-step prior is a useful alternative to shortcut extraction, especially for selected multimodal manipulation regimes.

\begin{figure}[t!]
    \centering
    \includegraphics[width=0.82\linewidth]{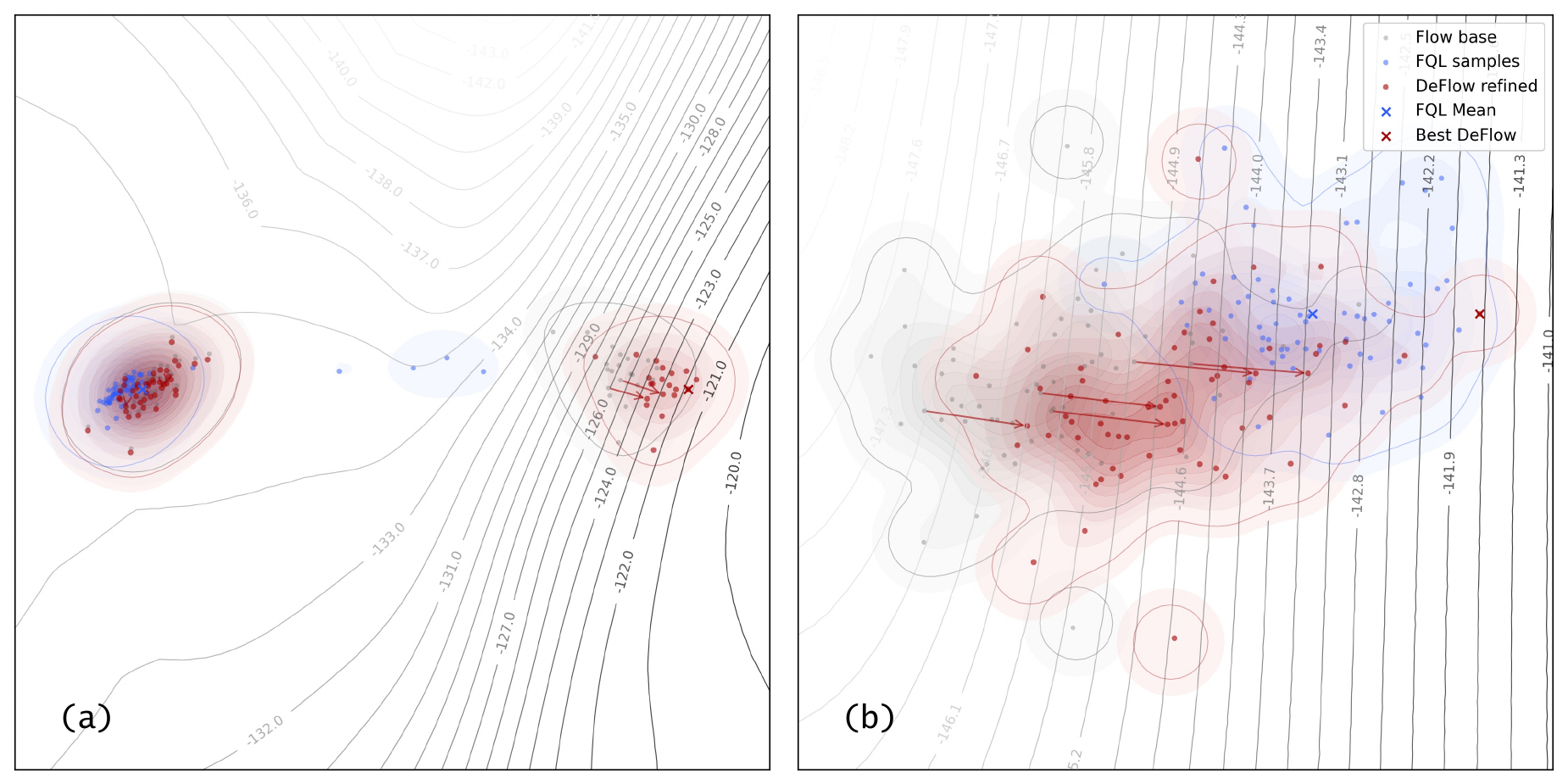}
    \caption{
   \textbf{Visualization of Action Landscapes.}
    Overlay of Q-value contours and action samples.
    \textcolor{gray}{\textbf{Gray}}: Base Flow samples (data proxy); \textcolor{blue}{\textbf{Blue}}: FQL (one-step); \textcolor{red}{\textbf{Red}}: DeFlow (ours). Arrows indicate the refinement trajectory.
    DeFlow performs local residual updates around base-flow proposals, illustrating smaller local shifts in this representative state.
    }
    \label{fig:policy_ext}
\end{figure}

\paragraph{Mechanism visualization.}
\Cref{fig:policy_ext} visualizes critical states in \texttt{cube-double-play}.
In these visualized states, FQL samples can fall between nearby modes or toward high-Q regions, while DeFlow starts from multi-step flow samples and applies a regularized residual.
This diagnostic is not a formal support guarantee, but it illustrates the intended role of the residual penalty: keep value improvement local to the learned proposal distribution.

\begin{figure}[t!]
    \centering
    \includegraphics[width=0.84\linewidth]{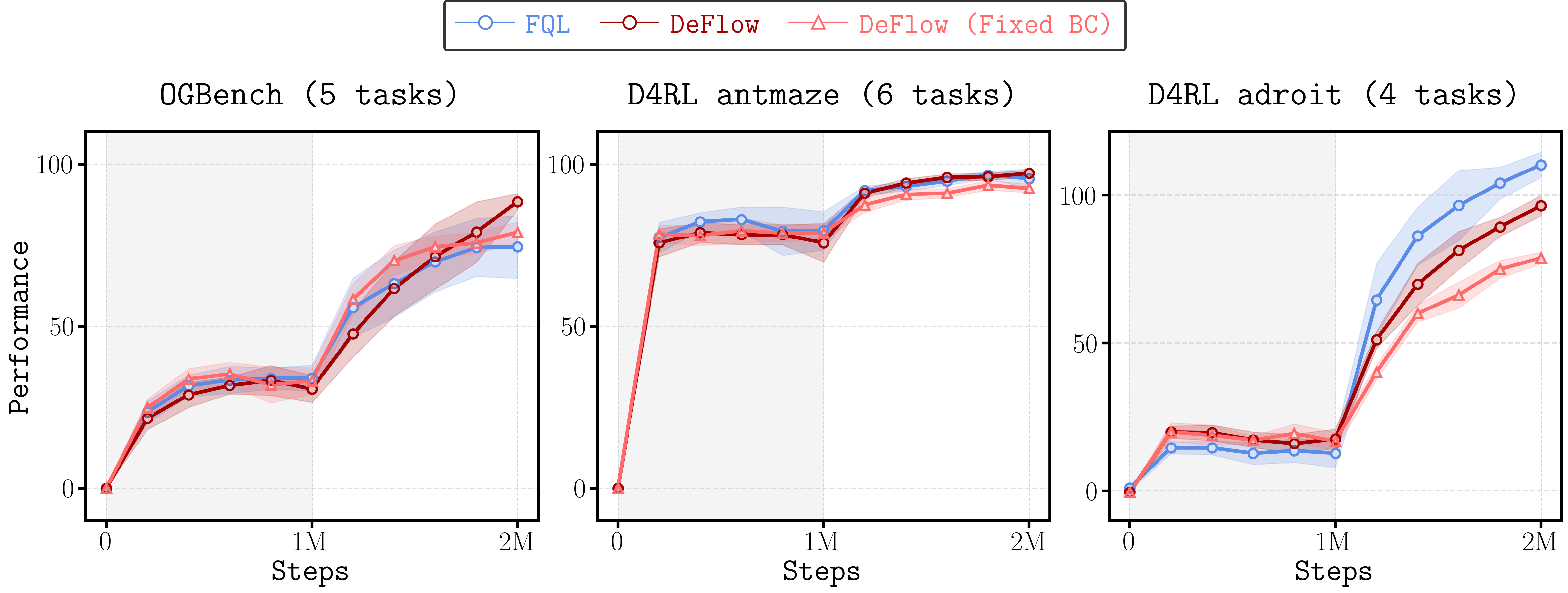}
    \caption{
    \textbf{Offline-to-online RL results ($\mathbf{8}$ seeds).}
    }
    \label{fig:o2o}
\end{figure}

\paragraph{Offline-to-online and diagnostics.}
\Cref{table:o2o_full,fig:o2o} summarize O2O results.
Using the same offline configuration, DeFlow is competitive with FQL on many tasks and is especially strong on \texttt{cube-double} and \texttt{puzzle-4x4}; it is weaker on some cloned Adroit tasks, highlighting dependence on prior coverage and critic quality.
Additional diagnostics in \cref{sec:diagnostics,sec:alpha} report $\delta$ sensitivity, an imbalanced human/expert dataset, action-conditioned refinement ablations, and compatibility with floq-style value modeling.
\Cref{sec:attribution,sec:capacity} add prior-only evaluation and residual-capacity sensitivity; \cref{app:runtime} documents the runtime comparison.

\subsection{Interpreting the Empirical Pattern}

The SORL comparison changes how the results should be read: DeFlow is not uniformly stronger, and the table points to regimes where different extraction mechanisms are useful.
SORL and shortcut-style methods can allocate extra evaluation compute across shortcut step counts or critic-ranked candidates, which fits several navigation and scene-style tasks.
The strong SORL numbers on \texttt{antmaze-large}, \texttt{humanoidmaze-medium}, \texttt{antsoccer}, and \texttt{scene} are consistent with this view.

DeFlow is intended for a narrower regime: the learned prior already produces useful local modes, but value extraction must make small, sample-specific corrections without collapsing those modes into a single shortcut output.
This is a possible explanation for the manipulation results, not an experimentally isolated cause.
For example, \texttt{cube-double} and \texttt{puzzle-3x3} require choosing among nearby multimodal behaviors and then making precise local adjustments.
A residual that is conditioned on both the state and the sampled action can move one proposal without forcing all proposals from the same state to share the same correction.
This is also why the landscape visualization is included as a diagnostic rather than as a proof: it illustrates the intended local mechanism in a representative state, while the quantitative table determines where that mechanism is actually useful.

The O2O results provide a complementary stress test: we reuse the same residual architecture and budget mechanism while appending online transitions to replay.
This transfers well when the offline prior already covers useful behaviors, as in \texttt{cube-double} and \texttt{puzzle-4x4}, but is weaker when prior coverage or critic quality is less reliable.
The appendix diagnostics probe sensitivity and component attribution rather than serving as additional headline benchmarks.

\paragraph{Attribution and residual capacity.}
Disabling the residual at the same trained checkpoints gives $10\pm2$ versus $62\pm12$ with refinement on cube-double task1, and $0.5\pm0.9$ versus $35\pm11$ on puzzle-3x3 task2 (\cref{tab:prior_only}). These evaluations isolate the residual's inference-time contribution conditional on the trained prior, not robustness to arbitrary prior quality. The capacity sweep (\cref{tab:capacity_core,tab:capacity_breadth}) also shows task dependence: one layer underperforms two on puzzle-3x3 and several navigation tasks, while greater depth does not consistently help on cube-double. A possible explanation is a balance between insufficient correction capacity and fitting optimistic critic errors. We have not directly measured this causal mechanism. The fixed four-layer main benchmark remains unchanged.

\subsection{Runtime and Compute Trade-offs}
\label{sec:run}

DeFlow separates behavior-prior sampling from value improvement.
The base policy still uses a multi-step flow solver, so DeFlow does not claim the same inference profile as one-step shortcut policies; the practical question is whether the residual module adds much overhead beyond the flow proposal itself.
Let $M$ denote the number of flow-solver steps, $C_v$ the cost of one vector-field evaluation, $C_f$ the cost of the residual module, and $C_Q$ the cost of one critic evaluation.
A direct BPTT actor update through an iterative generator retains the solver trajectory and backpropagates through $M$ vector-field calls, with memory proportional to the unrolled solver.
Shortcut policies reduce this to a one-step or few-step extraction cost, and SORL can spend test-time compute on shortcut step counts or critic-ranked candidates.
DeFlow still pays $M C_v$ to sample the prior, but its actor-gradient path only traverses $f_\phi$ and $Q$ because $a_{\text{flow}}=\operatorname{sg}[\mu_\psi(s,z)]$.

\begin{table}[H]
    \centering
    \caption{Compute trade-offs for generative policy extraction. $M$ is the number of flow-solver steps; $C_v$, $C_f$, and $C_Q$ denote vector-field, residual-module, and critic costs.}
    \label{tab:compute}
    \footnotesize
    \setlength{\tabcolsep}{5pt}
    \renewcommand{\arraystretch}{1.16}
    \begin{tabular*}{\linewidth}{@{}>{\bfseries}l@{\extracolsep{\fill}}lll@{}}
        \toprule
        Family & Update path & Eval. budget & Proposal use \\
        \midrule
        Iterative BPTT & solver + critic & $M C_v$ & update generator \\
        Shortcut / FQL & shortcut + critic & 1--few steps & compress actor \\
        SORL-style & shortcut + critic & $k$ steps; optional ranking & scale inference \\
        DeFlow & residual + critic & $M C_v + C_f$ & local residual \\
        \bottomrule
    \end{tabular*}
\end{table}

\Cref{tab:timing} reports wall-clock measurements on the same NVIDIA 3090 setup used for our main experiments.
The measured overhead is modest in this implementation: DeFlow is slightly slower than FQL in offline training and inference because it preserves multi-step proposal sampling, while the residual MLP adds little extra cost relative to the flow and critic updates.
These timings are implementation-specific and do not establish a speed advantage over SORL or a scaling law for larger models. DeFlow can also refine several proposals and rank them by the critic, at additional sampling and critic cost; that variant is not evaluated here. Applying the same separation to a large VLA-style prior is a motivation for future work, not an empirical result of this study.

\begin{table}[H]
    \centering
    \caption{Training and inference time comparison in milliseconds per step.}
    \label{tab:timing}
    \small
    \begin{tabular*}{\linewidth}{@{\extracolsep{\fill}}lccc}
        \toprule
        Method & \texttt{Offline train} & \texttt{Online train} & \texttt{Inference} \\
        \midrule
        \texttt{FQL} & $2.2$ {\tiny $\pm 1.0$} & $5.1$ {\tiny $\pm 0.5$} & $4.2$ {\tiny $\pm 0.7$} \\
        \texttt{DeFlow} & $2.4$ {\tiny $\pm 1.0$} & $5.3$ {\tiny $\pm 0.4$} & $4.5$ {\tiny $\pm 0.5$} \\
        \texttt{DeFlow (frozen prior)} & -- & $5.2$ {\tiny $\pm 0.5$} & -- \\
        \bottomrule
    \end{tabular*}
\end{table}

\section{Related Work}

\paragraph{Offline RL under behavior constraints.}
Offline RL often improves value under constraints that keep the policy close to the static dataset~\citep{levine2020offline}.
Conservative and behavior-regularized methods use value penalties or actor regularization~\citep{cql_kumar2020,td3bc_fujimoto2021}; IQL and SQL avoid explicit out-of-distribution maximization~\mbox{\citep{iql_kostrikov2022,sql_xu2023}}.
OGBench~\mbox{\citep{ogbench_park2025}} exposes a harder version of the same problem: the policy class must represent multimodal behavior, but the extraction step must still avoid critic exploitation.
DeFlow follows this constrained-improvement view with a residual budget around samples from a learned flow prior.

\paragraph{Diffusion and flow policies for offline RL.}
Diffusion policies and flow-matching policies are attractive for offline RL because they can model complex conditional action distributions.
Value guidance and weighted policy objectives offer several ways to incorporate critic information into expressive policies~\citep{qgpo_lu2023,edp_kang2023,qipo_zhang2025,awr_peng2019}.
Other methods backpropagate value gradients through the iterative denoising or sampling process~\citep{dql_wang2023,consistencyac_ding2024}, which provides a more direct policy-improvement signal but introduces substantial memory and stability costs.
Q-score matching~\citep{psenka2023learning} instead connects the policy score to the action gradient of the Q-function and differentiates through the denoising model rather than the full sampling chain. Thus, avoiding BPTT is an established design property; DeFlow's distinction is to apply value gradients to a proposal-conditioned action residual while training the prior only through behavior modeling.
Flow matching offers straighter generation paths and has recently been used for policy and value modeling in FQL, floq, SAC Flow, MeanFlow-QL, SORL, and Value Flows~\citep{flow_lipman2023,fql_park2025,agrawalla2025floq,zhang2025sacflow,wang2025one,espinosa2025scaling,dong2025valueflows}.
DeFlow is closest to this flow-based line, but keeps the iterative flow model as a behavior prior and applies critic gradients only to a small residual extractor.

\paragraph{Shortcut extraction and SORL.}
FQL and related shortcut methods reduce the optimization burden by distilling or parameterizing a one-step policy, making direct Q optimization cheaper than differentiating through a full solver~\citep{fql_park2025,shortcut_frans2025}.
SORL extends this efficient extraction direction with step-size-conditioned shortcut models, shortcut-consistency training, and inference-time evaluation over shortcut step counts, with optional critic best-of-$N$ selection; its strong results on several OGBench navigation and scene tasks make it an important comparison point~\citep{espinosa2025scaling}.
DeFlow explores a different trade-off: retain a multi-step learned prior and regularize a local residual instead of using a shortcut policy.
The empirical pattern in \cref{table:offline} is consistent with different extraction mechanisms being useful in different regimes; the unmatched inference budgets do not isolate the cause of this pattern.

\paragraph{Residual policy learning and refinement.}
Residual learning adds learned corrections on top of existing models, from residual networks~\citep{he2016deep} to residual policy learning~\citep{silver2019residual}.
Recent work trains small refinement modules on larger policies, often with online interaction~\citep{ankile2024refinement,xiao2025selfimproving,dong2025expo}.
DeFlow uses the same broad decomposition in offline policy extraction, with an action-conditioned residual so different samples from the same state can receive different local corrections.

\Needspace{12\baselineskip}
\section{Discussions and Conclusion}
\label{sec:conclusion}
We presented DeFlow, a BPTT-free framework for extracting policies from multi-step flow priors through action-conditioned residual refinement.
The empirical picture is deliberately scoped: DeFlow is competitive with recent flow-based offline RL methods, improves selected multimodal manipulation and O2O tasks, and underperforms strong shortcut or inference-scaling methods in some navigation regimes.
This suggests that offline generative policy extraction should be evaluated along several axes at once: behavior-prior fidelity, local value improvement, inference-time compute, and robustness to critic error.

\paragraph{Limitations.}
DeFlow depends on coverage of useful proposals: a local residual has no reliable mechanism for recovering distant behaviors absent from the data or missed by the prior. Proximity to an imperfect prior does not certify true data support. Both the budget $\delta$ and residual capacity affect performance, and adaptive $\alpha$ does not remove these choices or prevent critic exploitation. Our attribution tests do not systematically vary prior quality, and the critic-optimism explanation remains a hypothesis. Multi-step sampling retains a cost proportional to the solver budget, and the SORL comparison is not compute-matched.

\section*{Acknowledgments and Disclosures}
Zhancun Mu thanks his former advisor, Yitao Liang, for his support.
This work was supported by the Shanghai Artificial Intelligence Innovation Center under Grant No. 2025SHZDZX026D22.
The authors declare no competing interests relevant to this work.


\bibliographystyle{plainnat}
\bibliography{main}

%%%%%%%%%%%%%%%%%%%%%%%%%%%%%%%%%%%%%%%%%%%%%%%%%%%%%%%%%%%%%%%%%%%%%%%%%%%%%%%
%%%%%%%%%%%%%%%%%%%%%%%%%%%%%%%%%%%%%%%%%%%%%%%%%%%%%%%%%%%%%%%%%%%%%%%%%%%%%%%
% APPENDIX
%%%%%%%%%%%%%%%%%%%%%%%%%%%%%%%%%%%%%%%%%%%%%%%%%%%%%%%%%%%%%%%%%%%%%%%%%%%%%%%
%%%%%%%%%%%%%%%%%%%%%%%%%%%%%%%%%%%%%%%%%%%%%%%%%%%%%%%%%%%%%%%%%%%%%%%%%%%%%%%
\clearpage
\appendix
\onecolumn

\section{Residual Trust Region as an OT Surrogate}
\label{app:ot}

We use the standard Wasserstein formulation of optimal transport, where a Monge map deterministically pushes one distribution to another and $W_2^2$ is the minimum squared Euclidean cost over all couplings~\citep{ot_villani2009,computational_ot_peyre2019}.

\begin{proposition}
Fix a state $s$ and let $\nu_s=\mu_\psi(\cdot|s)$ be the learned flow prior. Assume that $\nu_s$ and the residual have finite second moments.
Define the refined policy by the deterministic map $T_s(a_b)=a_b+f_\phi(s,a_b)$ with $a_b\sim\nu_s$, i.e., $\pi_\phi(\cdot|s)=(T_s)_\#\nu_s$.
Then
\begin{equation}
W_2^2\!\left(\pi_\phi(\cdot|s),\nu_s\right)
\le
\mathbb{E}_{a_b\sim\nu_s}\left[\|f_\phi(s,a_b)\|_2^2\right].
\end{equation}
\end{proposition}

\begin{proof}
The joint distribution induced by $(a_b,T_s(a_b))$ is a valid coupling between $\nu_s$ and $\pi_\phi(\cdot|s)$.
The squared transport cost under this coupling is
$\mathbb{E}_{a_b\sim\nu_s}\|a_b-T_s(a_b)\|_2^2
=\mathbb{E}_{a_b\sim\nu_s}\|f_\phi(s,a_b)\|_2^2$.
Since $W_2^2$ is the minimum over all couplings, the inequality follows.
\end{proof}

This proposition only relates the refined policy to the learned flow prior.
It does not imply exact preservation of the true data support, which depends on the quality and coverage of $\mu_\psi$.
With the implementation's coordinate-averaged penalty, the upper bound is $d\,\mathbb{E}\|f_\phi(s,a_b)\|_2^2/d$. Averaging over replay states gives $\mathbb{E}_s W_2^2\le d\delta$ if the population constraint is satisfied. This is an average-state statement, not a per-state guarantee; a minibatch estimate alone is not a certified population bound. For bounded actions, the base proposal is clipped before refinement. Projection of the refined action onto the same box cannot increase its distance to that base proposal, so the same bound holds after action clipping.

\section{Implementation Details}
\label{app:implementation}

We implement DeFlow based on the FQL~\citep{fql_park2025} and floq~\citep{agrawalla2025floq} codebases. The flow model architecture follows the same design as FQL, using a $4$-layer MLP with $512$ hidden units. The refinement module uses the same architecture for clear comparison.
We use Q-Normalization as in FQL to stabilize Q-value scales across different tasks.
For other hyperparameters except the target divergence $\delta$, we follow the FQL/floq's default settings.
The experiments use public OGBench and D4RL benchmarks and the FQL/floq implementation bases under their respective terms and licenses.

\subsection{Hyperparameter Settings}
\label{sec:hyper}

For the target divergence $\delta$, we set it based on the estimated Intrinsic Action Variance (IAV) of the dataset. Specifically, we compute the average variance of actions among $k$-nearest neighbors ($k=5$) for each state in the following tasks and normalize it by the action dimension:

\begin{table}[H]
    \centering
    % \caption{Success rates on manipulation tasks. We report mean and standard deviation.}
    \label{tab:variance}
        \begin{tabular*}{\linewidth}{@{\extracolsep{\fill}}lcccc}
            \toprule
             & \texttt{cube-double} & \texttt{antmaze-large} & \texttt{puzzle-3x3} & \texttt{antsoccer} \\
            \midrule
            \texttt{IAV} & $0.013$ & $0.018$  & $0.007$ & $0.026$ \\
            \bottomrule
        \end{tabular*}
\end{table}

Leveraging prior knowledge of task complexity~\citep{ogbench_park2025}, we use the Intrinsic Action Variance (IAV) as a rough guide for choosing the constraint $\delta$:
\begin{itemize}
    \item \textbf{Fine Manipulation} (e.g., \texttt{cube-double-play}): High precision requires a tighter constraint, $\delta \approx 0.1 \times \text{IAV}$.
    \item \textbf{Navigation/Locomotion} (e.g., \texttt{antsoccer}): Higher tolerance allows a looser constraint, $\delta \approx 1.0 \times \text{IAV}$.
    \item \textbf{D4RL Benchmarks}: A generic setting suffices based on data quality (e.g., $\delta = 10^{-3}$).
\end{itemize}
IAV is a rough scale heuristic, not a validated predictor of an optimal budget. The actual settings include exceptions: humanoidmaze uses $0.001$, puzzles use $0.0005$, D4RL AntMaze uses $0.015$, and pen uses $0.01$. The complete task table, rather than the generic heuristic above, specifies the reported configuration.
We maintain the same $\delta$ for online fine-tuning without further adjustment. This strict protocol may limit asymptotic performance on specific outlier tasks, but it tests whether the residual constraint can transfer without task-specific retuning.
Detailed hyperparameter settings for all experiments are provided in \cref{table:method_hyp}.
\paragraph{Selection protocol and reporting scope.}
\Cref{tab:hp_protocol} separates inherited settings, task-dependent choices, and post-submission diagnostics. The main benchmark used a four-layer residual throughout; additional depth comparisons do not replace those scores. We retain the same task budget during O2O adaptation. The original records do not provide a complete count of candidate trials or environment evaluations used during hyperparameter development. Thus, the reported configuration and diagnostic budgets are reproducible, but they should not be interpreted as a complete audit of historical selection cost.

\begin{table}[H]
\centering
\caption{Hyperparameter choices and diagnostic scope. The sensitivity grids are diagnostic settings, not a claimed search history for selecting the main results.}
\label{tab:hp_protocol}
\small
\renewcommand{\arraystretch}{1.12}
\begin{tabularx}{\linewidth}{@{}lXX@{}}
\toprule
Quantity & Main benchmark protocol & Additional analysis / budget \\
\midrule
Residual budget $\delta$ & Task-dependent IAV-guided choice; retained for O2O & $\{0.0005,0.001,0.005,0.01,0.02\}$ diagnostic in \cref{tab:delta_sensitivity} \\
Penalty $\alpha$ & Adapted in log space; not a fixed-coefficient grid & Original nominal fixed settings $\{0.03,0.1,0.3,1,3\}$ in \cref{tab:alpha} \\
Residual architecture & Four hidden layers, width 512; no per-task depth tuning & One/two layers, width 512; four seeds per setting; \cref{sec:capacity} \\
Prior and critic & FQL-based configuration; prior learns by flow matching & Held at the same architecture within each capacity comparison \\
Inference budget & Multi-step proposal followed by one residual & Ten Euler steps in capacity runs; no candidate ranking \\
Evaluation selection & Fixed-step main evaluation & Final scheduled checkpoint for capacity; 20 or 50 episodes per seed \\
\bottomrule
\end{tabularx}
\end{table}

\paragraph{Implementation conventions.}
The residual penalty is the minibatch mean of squared coordinates, $\widehat g=(Bd)^{-1}\sum_{i=1}^{B}\|\Delta a_i\|_2^2$. The actor treats $\alpha$ as constant, and the dual update treats $\widehat g$ as constant. In the audited adaptive implementation, $\eta=\log\alpha$ starts at 1 and uses the shared Adam learning rate $3\times10^{-4}$. Q-normalization divides the actor objective by the stopped minibatch mean absolute Q-value. Clipping is applied to the base proposal and the refined action. These conventions matter when translating a reported $\delta$ into an action-space displacement.

\section{Additional Experimental Results}

\label{sec:add}

We report the full offline RL results across all $73$ tasks in \cref{table:offline_full}. The results are averaged over 8 seeds (4 seeds for pixel-based tasks) and we bold scores that are within 95\% of the best performance as in FQL~\citep{fql_park2025}.

\subsection{Additional Diagnostics}
\label{sec:diagnostics}

This section collects diagnostics that are referenced in the main text but are not used as primary benchmark claims.
They are intended to probe failure modes and component attribution.

\begin{table}[H]
    \centering
    \caption{Sensitivity to the residual budget $\delta$ on a representative manipulation diagnostic. Larger budgets permit larger residual moves and can quickly degrade performance.}
    \label{tab:delta_sensitivity}
    \begin{tabular*}{\linewidth}{@{\extracolsep{\fill}}lccccc}
        \toprule
        $\delta$ & $0.0005$ & $0.001$ & $0.005$ & $0.01$ & $0.02$ \\
        \midrule
        Score & $55$ {\tiny $\pm 8$} & $50$ {\tiny $\pm 7$} & $27$ {\tiny $\pm 4$} & $8$ {\tiny $\pm 5$} & $4$ {\tiny $\pm 2$} \\
        \bottomrule
    \end{tabular*}
\end{table}

\begin{table}[H]
    \centering
    \caption{Diagnostic checks for prior coverage and action-conditioned refinement. The mixed dataset uses 50\% human and 50\% expert data; the refinement ablation removes action conditioning from $f_\phi(s,a_b)$.}
    \label{tab:diagnostic_ablation}
    \begin{tabular*}{\linewidth}{@{\extracolsep{\fill}}llcc}
        \toprule
        Diagnostic & Setting & Baseline & DeFlow / Variant \\
        \midrule
        Mixed human/expert data & score & FQL: $21$ {\tiny $\pm 3$} & DeFlow: $37$ {\tiny $\pm 5$} \\
        Refinement module & full action-conditioned $f_\phi(s,a_b)$ & - & $40$ {\tiny $\pm 5$} \\
        Refinement module & remove action conditioning & - & $5$ {\tiny $\pm 3$} \\
        \bottomrule
    \end{tabular*}
\end{table}

\begin{table}[H]
    \centering
    \caption{Success rates on manipulation tasks when combining DeFlow with floq's value function.}
    \label{tab:floq}
        \begin{tabular*}{\linewidth}{@{\extracolsep{\fill}}lcccc}
            \toprule
            \texttt{Task} & \texttt{FQL} & \texttt{floq} & \texttt{DeFlow} & \texttt{DeFlow+floq} \\
            \midrule
            \texttt{cube-double-play} & $25$ {\tiny $\pm 6$} & $47$ {\tiny $\pm 15$} & $40$ {\tiny $\pm 5$} & $\mathbf{58}$ {\tiny $\pm 5$} \\
            \texttt{puzzle-3x3-play}  & $29$ {\tiny $\pm 5$} & $37$ {\tiny $\pm 7$}  & $43$ {\tiny $\pm 4$} & $\mathbf{50}$ {\tiny $\pm 3$} \\
            \bottomrule
        \end{tabular*}
\end{table}

\subsection{Ablations on \texorpdfstring{$\alpha$}{alpha} Adaptation}
\label{sec:alpha}

We retain the original comparison of adaptive penalties with fixed-coefficient settings on \texttt{cube-double-play} and \texttt{puzzle-3x3-play}, with the implementation qualification below.

\begin{table}[H]
    \centering
    \caption{Original fixed-penalty sweep versus adaptation, separated by training phase. The fixed columns retain the nominal sweep setting $c$; see the implementation qualification below.}
    \label{tab:alpha}
    \footnotesize
    \setlength{\tabcolsep}{2pt}
    \begin{tabular*}{\linewidth}{@{\extracolsep{\fill}}llcccccc@{}}
            \toprule
             Task & Phase & $c=0.03$ & $0.1$ & $0.3$ & $1$ & $3$ & Adaptive \\
            \midrule
             cube-double & Offline & $0\pm0$ & $0\pm0$ & $0\pm0$ & $44\pm2$ & $7\pm2$ & $48\pm13$ \\
                         & Online & $0\pm0$ & $0\pm0$ & $0\pm0$ & $10\pm3$ & $89\pm5$ & $93\pm4$ \\
             \midrule
             puzzle-3x3 & Offline & $0\pm0$ & $0\pm0$ & $7\pm7$ & $0\pm0$ & $35\pm8$ & $23\pm8$ \\
                        & Online & $100\pm1$ & $100\pm0$ & $100\pm0$ & $0\pm0$ & $0\pm0$ & $100\pm0$ \\
            \bottomrule
    \end{tabular*}
\end{table}

The original sweep illustrates task- and phase-dependent sensitivity: small nominal settings perform poorly offline on cube-double, whereas a setting that works offline on puzzle-3x3 can fail online. Adaptation is competitive in these examples but does not win every offline comparison. These returns do not by themselves establish that failures are caused by OOD actions or that the adaptive constraint is always satisfied.

\paragraph{Implementation qualification.}
The current fixed-penalty code path multiplies the configured coefficient twice, giving an effective coefficient $c^2$. The historical code version for this original sweep has not been recovered, so we retain its nominal labels rather than assert a calibrated effective-$\alpha$ axis. This table is descriptive evidence about the reported settings, not a matched-effective-penalty comparison. The new capacity diagnostics all use the adaptive branch and are unaffected by that fixed-branch issue.

\subsection{Prior-Only and Residual Attribution}
\label{sec:attribution}

We evaluate the same four trained checkpoints per task with and without the residual at inference, using 50 episodes per checkpoint and mode. No component is retrained. Scores in \cref{tab:prior_only} are success percentages, reported as mean and standard deviation across checkpoints. This tests whether the residual changes performance for the particular learned priors; it does not isolate the effects of prior architecture, prior training quality, or dataset coverage.

\begin{table}[H]
\centering
\caption{Inference-time residual attribution at the same trained checkpoints. These are separate evaluations, not replacements for the main benchmark scores.}
\label{tab:prior_only}
\begin{tabular*}{\linewidth}{@{\extracolsep{\fill}}lcc@{}}
\toprule
Task & Base flow only & Base flow + residual \\
\midrule
cube-double task1 & $10\pm2$ & $62\pm12$ \\
puzzle-3x3 task2 & $0.5\pm0.9$ & $35\pm11$ \\
\bottomrule
\end{tabular*}
\end{table}

The improvement shows that these scores do not arise from the flow prior alone. It is complementary to the action-input ablation in \cref{tab:diagnostic_ablation}: disabling the residual tests its inference-time contribution, whereas removing action conditioning changes the correction function being learned. Neither establishes a guarantee of mode preservation. The residual is conditioned on the proposal's position in action space, without explicit mode labels.

\subsection{Residual-Capacity Sensitivity}
\label{sec:capacity}

We vary only the residual hidden-layer configuration within each new comparison, holding the flow prior and critic at four hidden layers of width 512. Each configuration uses four seeds, a fixed task budget, ten flow steps, batch size 256, and learning rate $3\times10^{-4}$. We report the last scheduled evaluation: one million offline updates for OGBench and 500,000 for pen-cloned. Cube-double uses seeds 2026--2029 and 20 evaluation episodes per seed; the other capacity runs use seeds 1996--1999 and 50 episodes. Means and population standard deviations are computed across the four final seed scores and rounded to whole points.

\begin{table}[H]
\centering
\caption{Residual-depth diagnostics. All hidden layers have width 512. Cube's four-layer entry is a matched four-seed rerun ($\dagger$); puzzle and scene four-layer entries are the original eight-seed benchmark references ($*$), not matched reruns.}
\label{tab:capacity_core}
\begin{tabular*}{\linewidth}{@{\extracolsep{\fill}}lccc@{}}
\toprule
Task & One layer & Two layers & Four layers \\
\midrule
cube-double task1 & $81\pm9$ & $74\pm23$ & $69\pm10\,\dagger$ \\
puzzle-3x3 task2 & $6\pm3$ & $55\pm7$ & $38\pm11\,*$ \\
scene task2 & $52\pm12$ & $70\pm13$ & $59\pm17\,*$ \\
\bottomrule
\end{tabular*}
\end{table}

The original cube-double task1 main-table result is $65\pm6$ across eight seeds; it is unchanged. The four-seed cube rerun above is distinct. Across these tasks, the results do not support a monotonic benefit from increasing residual depth. Comparisons to the original puzzle and scene references also differ in seeds and evaluation provenance, so they should not be treated as paired estimates of a depth effect.

\begin{table}[H]
\centering
\caption{Broader one- versus two-layer sensitivity diagnostic. OGBench entries are success percentages; pen-cloned uses D4RL normalized return. No best-depth selection is applied to the main benchmark.}
\label{tab:capacity_breadth}
\begin{tabular*}{\linewidth}{@{\extracolsep{\fill}}lcc@{}}
\toprule
Task & One layer & Two layers \\
\midrule
antmaze-large task2 & $48\pm14$ & $73\pm6$ \\
humanoidmaze-medium task2 & $43\pm18$ & $73\pm10$ \\
antsoccer-arena task2 & $28\pm12$ & $76\pm3$ \\
cube-single task2 & $98\pm1$ & $97\pm3$ \\
puzzle-4x4 task1 & $22\pm2$ & $24\pm7$ \\
pen-cloned & $63\pm20$ & $64\pm8$ \\
\bottomrule
\end{tabular*}
\end{table}

\paragraph{Interpretation and limits.}
Two layers improve over one on several navigation tasks and puzzle-3x3, whereas cube-single is nearly saturated and the pen-cloned and puzzle-4x4 differences are small relative to their variability. A plausible explanation is a trade-off between representational underfitting and optimization against an imperfect offline critic~\citep{levine2020offline}. Insufficient capacity may miss useful proposal-dependent corrections. Greater capacity may fit sharper directions of optimistic Q-error under a DDPG-style actor objective; a smaller network could sometimes provide useful implicit regularization. These experiments do not measure Q-calibration, residual smoothness, or optimization conditioning directly, so they do not establish that explanation or show that smaller networks are intrinsically smoother. The same $\delta$ sets the same target penalty budget, but actual residual norms can differ because the constraint is only encouraged during training.

\begin{table}[H]
\centering
\caption{Task-specific settings for the new capacity diagnostics. Target aggregation refers to the Bellman target, not candidate ranking.}
\label{tab:capacity_protocol}
\small
\begin{tabular*}{\linewidth}{@{\extracolsep{\fill}}lccc@{}}
\toprule
Task & $\delta$ & Discount & Target aggregation \\
\midrule
cube-double task1 & $0.001$ & $0.99$ & mean \\
puzzle-3x3 task2 & $0.0005$ & $0.99$ & mean \\
scene task2 & $0.001$ & $0.99$ & mean \\
antmaze-large task2 & $0.01$ & $0.99$ & min \\
humanoidmaze-medium task2 & $0.001$ & $0.995$ & mean \\
antsoccer-arena task2 & $0.01$ & $0.995$ & mean \\
cube-single task2 & $0.001$ & $0.99$ & mean \\
puzzle-4x4 task1 & $0.0005$ & $0.99$ & mean \\
pen-cloned & $0.01$ & $0.99$ & min \\
\bottomrule
\end{tabular*}
\end{table}

\section{Runtime and Compute Accounting}
\label{app:runtime}
\Cref{tab:timing} reports the original NVIDIA RTX 3090 per-step timings. FQL and DeFlow take $2.2\pm1.0$ and $2.4\pm1.0$ ms per offline training step, respectively, and $4.2\pm0.7$ and $4.5\pm0.5$ ms per inference step. DeFlow's online training time is $5.3\pm0.4$ ms, versus $5.2\pm0.5$ ms when freezing the prior. These measurements characterize this implementation, not a controlled decomposition of residual-head overhead or an equal-compute comparison with SORL.

The additional capacity runs use ten Euler flow steps, batch size 256, and one million training updates for each OGBench run (500,000 for pen-cloned). Evaluation counts, seeds, and task-specific settings are listed in \cref{sec:capacity,tab:capacity_protocol}. These update counts exclude evaluation and compilation time. The original timing record does not specify enough detail about warm-up, synchronization, timing repeats, or evaluation overhead to infer total experimental GPU-hours from per-step timings. We therefore do not report an extrapolated total-compute estimate.

The architecture avoids an actor gradient through the flow solver but still pays for its forward evaluations. A larger pretrained action prior could make this separation attractive, including VLA-style priors, but neither VLA adaptation nor large-model runtime scaling is tested here.

\clearpage

\begin{table}[H]
\caption{
\textbf{Task-specific hyperparameters for offline RL.} 
We take hyperparameters from FQL~\citep{fql_park2025} for comparison.
}
\label{table:method_hyp}
\centering
\setlength{\tabcolsep}{2.5pt}
\resizebox{\textwidth}{!}{%
\newlength{\mylen}
\setlength{\mylen}{19pt}
\begin{tabular}{l@{\hskip \mylen}c@{\hskip \mylen}c@{\hskip \mylen}c@{\hskip \mylen}c@{\hskip \mylen}c@{\hskip \mylen}c@{\hskip \mylen}c@{\hskip \mylen}c@{\hskip \mylen}c@{\hskip \mylen}c}
\toprule
& \texttt{IQL} & \texttt{ReBRAC} & \texttt{IDQL} & \texttt{SRPO} & \texttt{CAC} & \texttt{FAWAC} & \texttt{FBRAC} & \texttt{IFQL} & \texttt{\color{myblue}FQL} & \texttt{\color{myred}DeFlow} \\
\texttt{Task} & $\alpha$ & $(\alpha_1, \alpha_2)$ & $N$ & $\beta$ & $\eta$ & $\alpha$ & $\alpha$ & $N$ & {\color{myblue}$\alpha$} & {\color{myred}$\delta$} \\
\midrule

\texttt{antmaze-large-navigate-singletask-task1-v0} & $10$ & $(0.003, 0.01)$ & $32$ & $0.3$ & $1$ & $3$ & $3$ & $32$ & $10$ & $0.01$\\
\texttt{antmaze-large-navigate-singletask-task2-v0} & $10$ & $(0.003, 0.01)$ & $32$ & $0.3$ & $1$ & $3$ & $3$ & $32$ & $10$ & $0.01$ \\
\texttt{antmaze-large-navigate-singletask-task3-v0} & $10$ & $(0.003, 0.01)$ & $32$ & $0.3$ & $1$ & $3$ & $3$ & $32$ & $10$ & $0.01$ \\
\texttt{antmaze-large-navigate-singletask-task4-v0} & $10$ & $(0.003, 0.01)$ & $32$ & $0.3$ & $1$ & $3$ & $3$ & $32$ & $10$ & $0.01$ \\
\texttt{antmaze-large-navigate-singletask-task5-v0} & $10$ & $(0.003, 0.01)$ & $32$ & $0.3$ & $1$ & $3$ & $3$ & $32$ & $10$ & $0.01$ \\
\midrule
\texttt{antmaze-giant-navigate-singletask-task1-v0} & $10$ & $(0.003, 0.01)$ & $32$ & $0.3$ & $1$ & $3$ & $10$ & $32$ & $10$ & $0.01$ \\
\texttt{antmaze-giant-navigate-singletask-task2-v0} & $10$ & $(0.003, 0.01)$ & $32$ & $0.3$ & $1$ & $3$ & $10$ & $32$ & $10$ & $0.01$ \\
\texttt{antmaze-giant-navigate-singletask-task3-v0} & $10$ & $(0.003, 0.01)$ & $32$ & $0.3$ & $1$ & $3$ & $10$ & $32$ & $10$ & $0.01$ \\
\texttt{antmaze-giant-navigate-singletask-task4-v0} & $10$ & $(0.003, 0.01)$ & $32$ & $0.3$ & $1$ & $3$ & $10$ & $32$ & $10$ & $0.01$ \\
\texttt{antmaze-giant-navigate-singletask-task5-v0} & $10$ & $(0.003, 0.01)$ & $32$ & $0.3$ & $1$ & $3$ & $10$ & $32$ & $10$ & $0.01$ \\
\midrule
\texttt{humanoidmaze-medium-navigate-singletask-task1-v0} & $10$ & $(0.01, 0.01)$ & $32$ & $0.3$ & $0.03$ & $3$ & $30$ & $32$ & $30$ & $0.001$ \\
\texttt{humanoidmaze-medium-navigate-singletask-task2-v0} & $10$ & $(0.01, 0.01)$ & $32$ & $0.3$ & $0.03$ & $3$ & $30$ & $32$ & $30$ & $0.001$ \\
\texttt{humanoidmaze-medium-navigate-singletask-task3-v0} & $10$ & $(0.01, 0.01)$ & $32$ & $0.3$ & $0.03$ & $3$ & $30$ & $32$ & $30$ & $0.001$ \\
\texttt{humanoidmaze-medium-navigate-singletask-task4-v0} & $10$ & $(0.01, 0.01)$ & $32$ & $0.3$ & $0.03$ & $3$ & $30$ & $32$ & $30$ & $0.001$ \\
\texttt{humanoidmaze-medium-navigate-singletask-task5-v0} & $10$ & $(0.01, 0.01)$ & $32$ & $0.3$ & $0.03$ & $3$ & $30$ & $32$ & $30$ & $0.001$ \\
\midrule
\texttt{humanoidmaze-large-navigate-singletask-task1-v0} & $10$ & $(0.01, 0.01)$ & $32$ & $0.3$ & $1$ & $3$ & $30$ & $32$ & $30$ & $0.001$\\
\texttt{humanoidmaze-large-navigate-singletask-task2-v0} & $10$ & $(0.01, 0.01)$ & $32$ & $0.3$ & $1$ & $3$ & $30$ & $32$ & $30$ & $0.001$\\
\texttt{humanoidmaze-large-navigate-singletask-task3-v0} & $10$ & $(0.01, 0.01)$ & $32$ & $0.3$ & $1$ & $3$ & $30$ & $32$ & $30$ & $0.001$\\
\texttt{humanoidmaze-large-navigate-singletask-task4-v0} & $10$ & $(0.01, 0.01)$ & $32$ & $0.3$ & $1$ & $3$ & $30$ & $32$ & $30$ & $0.001$\\
\texttt{humanoidmaze-large-navigate-singletask-task5-v0} & $10$ & $(0.01, 0.01)$ & $32$ & $0.3$ & $1$ & $3$ & $30$ & $32$ & $30$ & $0.001$\\
\midrule
\texttt{antsoccer-arena-navigate-singletask-task1-v0} & $1$ & $(0.01, 0.01)$ & $32$ & $0.03$ & $1$ & $10$ & $30$ & $64$ & $10$ &$0.01$\\
\texttt{antsoccer-arena-navigate-singletask-task2-v0} & $1$ & $(0.01, 0.01)$ & $32$ & $0.03$ & $1$ & $10$ & $30$ & $64$ & $10$ &$0.01$\\
\texttt{antsoccer-arena-navigate-singletask-task3-v0} & $1$ & $(0.01, 0.01)$ & $32$ & $0.03$ & $1$ & $10$ & $30$ & $64$ & $10$ &$0.01$\\
\texttt{antsoccer-arena-navigate-singletask-task4-v0} & $1$ & $(0.01, 0.01)$ & $32$ & $0.03$ & $1$ & $10$ & $30$ & $64$ & $10$ &$0.01$\\
\texttt{antsoccer-arena-navigate-singletask-task5-v0} & $1$ & $(0.01, 0.01)$ & $32$ & $0.03$ & $1$ & $10$ & $30$ & $64$ & $10$ &$0.01$\\
\midrule
\texttt{cube-single-play-singletask-task1-v0} & $1$ & $(1, 0)$ & $32$ & $0.03$ & $0.003$ & $1$ & $100$ & $32$ & $300$ & $0.001$ \\
\texttt{cube-single-play-singletask-task2-v0} & $1$ & $(1, 0)$ & $32$ & $0.03$ & $0.003$ & $1$ & $100$ & $32$ & $300$ & $0.001$ \\
\texttt{cube-single-play-singletask-task3-v0} & $1$ & $(1, 0)$ & $32$ & $0.03$ & $0.003$ & $1$ & $100$ & $32$ & $300$&  $0.001$\\
\texttt{cube-single-play-singletask-task4-v0} & $1$ & $(1, 0)$ & $32$ & $0.03$ & $0.003$ & $1$ & $100$ & $32$ & $300$ & $0.001$\\
\texttt{cube-single-play-singletask-task5-v0} & $1$ & $(1, 0)$ & $32$ & $0.03$ & $0.003$ & $1$ & $100$ & $32$ & $300$ & $0.001$\\
\bottomrule
\end{tabular}
}
\end{table}
\clearpage
\begin{table}[H]
\ContinuedFloat
\caption{\textbf{Task-specific hyperparameters for offline RL (continued).}}
\centering
\setlength{\tabcolsep}{2.5pt}
\resizebox{\textwidth}{!}{%
\setlength{\mylen}{19pt}
\begin{tabular}{l@{\hskip \mylen}c@{\hskip \mylen}c@{\hskip \mylen}c@{\hskip \mylen}c@{\hskip \mylen}c@{\hskip \mylen}c@{\hskip \mylen}c@{\hskip \mylen}c@{\hskip \mylen}c@{\hskip \mylen}c}
\toprule
& \texttt{IQL} & \texttt{ReBRAC} & \texttt{IDQL} & \texttt{SRPO} & \texttt{CAC} & \texttt{FAWAC} & \texttt{FBRAC} & \texttt{IFQL} & \texttt{\color{myblue}FQL} & \texttt{\color{myred}DeFlow} \\
\texttt{Task} & $\alpha$ & $(\alpha_1, \alpha_2)$ & $N$ & $\beta$ & $\eta$ & $\alpha$ & $\alpha$ & $N$ & {\color{myblue}$\alpha$} & {\color{myred}$\delta$} \\
\midrule

\texttt{cube-double-play-singletask-task1-v0} & $0.3$ & $(0.1, 0)$ & $32$ & $0.1$ & $0.3$ & $0.3$ & $100$ & $32$ & $300$ & $0.001$ \\
\texttt{cube-double-play-singletask-task2-v0} & $0.3$ & $(0.1, 0)$ & $32$ & $0.1$ & $0.3$ & $0.3$ & $100$ & $32$ & $300$ & $0.001$ \\
\texttt{cube-double-play-singletask-task3-v0} & $0.3$ & $(0.1, 0)$ & $32$ & $0.1$ & $0.3$ & $0.3$ & $100$ & $32$ & $300$ & $0.001$\\
\texttt{cube-double-play-singletask-task4-v0} & $0.3$ & $(0.1, 0)$ & $32$ & $0.1$ & $0.3$ & $0.3$ & $100$ & $32$ & $300$ & $0.001$\\
\texttt{cube-double-play-singletask-task5-v0} & $0.3$ & $(0.1, 0)$ & $32$ & $0.1$ & $0.3$ & $0.3$ & $100$ & $32$ & $300$ & $0.001$\\
\midrule
\texttt{scene-play-singletask-task1-v0} & $10$ & $(0.1, 0.01)$ & $32$ & $0.1$ & $0.3$ & $0.3$ & $100$ & $32$ & $300$ & $0.001$ \\
\texttt{scene-play-singletask-task2-v0} & $10$ & $(0.1, 0.01)$ & $32$ & $0.1$ & $0.3$ & $0.3$ & $100$ & $32$ & $300$ & $0.001$ \\
\texttt{scene-play-singletask-task3-v0} & $10$ & $(0.1, 0.01)$ & $32$ & $0.1$ & $0.3$ & $0.3$ & $100$ & $32$ & $300$ & $0.001$ \\
\texttt{scene-play-singletask-task4-v0} & $10$ & $(0.1, 0.01)$ & $32$ & $0.1$ & $0.3$ & $0.3$ & $100$ & $32$ & $300$ & $0.001$ \\
\texttt{scene-play-singletask-task5-v0} & $10$ & $(0.1, 0.01)$ & $32$ & $0.1$ & $0.3$ & $0.3$ & $100$ & $32$ & $300$ & $0.001$ \\
\midrule
\texttt{puzzle-3x3-play-singletask-task1-v0} & $10$ & $(0.3, 0.01)$ & $32$ & $0.1$ & $0.01$ & $0.3$ & $100$ & $32$ & $1000$ & $0.0005$\\
\texttt{puzzle-3x3-play-singletask-task2-v0} & $10$ & $(0.3, 0.01)$ & $32$ & $0.1$ & $0.01$ & $0.3$ & $100$ & $32$ & $1000$ & $0.0005$\\
\texttt{puzzle-3x3-play-singletask-task3-v0} & $10$ & $(0.3, 0.01)$ & $32$ & $0.1$ & $0.01$ & $0.3$ & $100$ & $32$ & $1000$ & $0.0005$\\
\texttt{puzzle-3x3-play-singletask-task4-v0} & $10$ & $(0.3, 0.01)$ & $32$ & $0.1$ & $0.01$ & $0.3$ & $100$ & $32$ & $1000$ & $0.0005$\\
\texttt{puzzle-3x3-play-singletask-task5-v0} & $10$ & $(0.3, 0.01)$ & $32$ & $0.1$ & $0.01$ & $0.3$ & $100$ & $32$ & $1000$ & $0.0005$\\
\midrule
\texttt{puzzle-4x4-play-singletask-task1-v0} & $3$ & $(0.3, 0.01)$ & $32$ & $0.1$ & $0.01$ & $0.3$ & $300$ & $32$ & $1000$ & $0.0005$ \\
\texttt{puzzle-4x4-play-singletask-task2-v0} & $3$ & $(0.3, 0.01)$ & $32$ & $0.1$ & $0.01$ & $0.3$ & $300$ & $32$ & $1000$ & $0.0005$ \\
\texttt{puzzle-4x4-play-singletask-task3-v0} & $3$ & $(0.3, 0.01)$ & $32$ & $0.1$ & $0.01$ & $0.3$ & $300$ & $32$ & $1000$ & $0.0005$ \\
\texttt{puzzle-4x4-play-singletask-task4-v0} & $3$ & $(0.3, 0.01)$ & $32$ & $0.1$ & $0.01$ & $0.3$ & $300$ & $32$ & $1000$ & $0.0005$ \\
\texttt{puzzle-4x4-play-singletask-task5-v0} & $3$ & $(0.3, 0.01)$ & $32$ & $0.1$ & $0.01$ & $0.3$ & $300$ & $32$ & $1000$ & $0.0005$ \\
\midrule
\texttt{antmaze-umaze-v2} & - & - & - & - & $0.01$ & $3$ & $10$ & $32$ & $10$ & $0.015$\\
\texttt{antmaze-umaze-diverse-v2} & - & - & - & - & $0.01$ & $3$ & $10$ & $32$ & $10$ & $0.015$\\
\texttt{antmaze-medium-play-v2} & - & - & - & - & $0.01$ & $3$ & $10$ & $32$ & $10$ & $0.015$ \\
\texttt{antmaze-medium-diverse-v2} & - & - & - & - & $0.01$ & $3$ & $10$ & $32$ & $10$ & $0.015$ \\
\texttt{antmaze-large-play-v2} & - & - & - & - & $4.5$ & $3$ & $1$ & $32$ & $3$ & $0.015$ \\
\texttt{antmaze-large-diverse-v2} & - & - & - & - & $3.5$ & $3$ & $1$ & $32$ & $3$ & $0.015$ \\
\midrule
\texttt{pen-human-v1} & - & - & $32$ & $0.03$ & $0.003$ & $0.03$ & $30000$ & $32$ & $10000$ & $0.01$ \\
\texttt{pen-cloned-v1} & - & - & $32$ & $0.1$ & $0.003$ & $0.3$ & $10000$ & $32$ & $10000$ & $0.01$ \\
\texttt{pen-expert-v1} & - & - & $32$ & $0.1$ & $0.03$ & $0.1$ & $30000$ & $32$ & $3000$ & $0.01$\\
\texttt{door-human-v1} & - & - & $32$ & $0.01$ & $0.03$ & $1$ & $30000$ & $32$ & $30000$ & $0.001$\\
\texttt{door-cloned-v1} & - & - & $32$ & $0.03$ & $0.03$ & $1$ & $10000$ & $128$ & $30000$ & $0.001$\\
\texttt{door-expert-v1} & - & - & $32$ & $0.01$ & $0.03$ & $3$ & $30000$ & $32$ & $30000$ & $0.001$\\
\texttt{hammer-human-v1} & - & - & $128$ & $0.1$ & $0.03$ & $3$ & $30000$ & $32$ & $30000$ & $0.001$\\
\texttt{hammer-cloned-v1} & - & - & $32$ & $0.1$ & $0.003$ & $0.03$ & $10000$ & $32$ & $10000$ & $0.001$\\
\texttt{hammer-expert-v1} & - & - & $32$ & $0.03$ & $0.03$ & $3$ & $30000$ & $32$ & $30000$ & $0.001$\\
\texttt{relocate-human-v1} & - & - & $32$ & $0.03$ & $0.01$ & $0.3$ & $30000$ & $128$ & $10000$ & $0.001$\\
\texttt{relocate-cloned-v1} & - & - & $64$ & $0.03$ & $0.01$ & $0.1$ & $3000$ & $32$ & $30000$ & $0.001$\\
\texttt{relocate-expert-v1} & - & - & $32$ & $0.01$ & $0.003$ & $1$ & $30000$ & $32$ & $30000$ & $0.001$\\
\midrule
\texttt{visual-cube-single-play-singletask-task1-v0} & $1$ & $(1, 0)$ & - & - & - & - & $100$ & $32$ & $300$ & $0.001$\\
\texttt{visual-cube-double-play-singletask-task1-v0} & $0.3$ & $(0.1, 0)$ & - & - & - & - & $100$ & $32$ & $100$ & $0.001$\\
\texttt{visual-scene-play-singletask-task1-v0} & $10$ & $(0.1, 0.01)$ & - & - & - & - & $100$ & $32$ & $100$ & $0.001$\\
\texttt{visual-puzzle-3x3-play-singletask-task1-v0} & $10$ & $(0.3, 0.01)$ & - & - & - & - & $100$ & $32$ & $300$ & $0.0005$ \\
\texttt{visual-puzzle-4x4-play-singletask-task1-v0} & $3$ & $(0.3, 0.01)$ & - & - & - & - & $300$ & $32$ & $300$ & $0.0005$\\

\bottomrule

\end{tabular}
}
\end{table}

\clearpage

\begin{table}[H]
\caption{
\textbf{Full offline RL results.}
We present the full results on the $73$ OGBench and D4RL tasks. \texttt{(*)} indicates the default task in each environment.
The results are averaged over $8$ seeds ($4$ seeds for pixel-based tasks) unless otherwise mentioned.
}
\label{table:offline_full}
\centering
\setlength{\tabcolsep}{2.5pt}
\resizebox{\textwidth}{!}{%
\begin{threeparttable}
\begin{tabular}{lccccccccccc}
\toprule
\multicolumn{1}{c}{} & \multicolumn{3}{c}{\texttt{Gaussian Policies}} & \multicolumn{3}{c}{\texttt{Diffusion Policies}} & \multicolumn{5}{c}{\texttt{Flow Policies}} \\
\cmidrule(lr){2-4} \cmidrule(lr){5-7} \cmidrule(lr){8-12}
\texttt{Task} & \texttt{BC} & \texttt{IQL} & \texttt{ReBRAC} & \texttt{IDQL} & \texttt{SRPO} & \texttt{CAC} & \texttt{FAWAC} & \texttt{FBRAC} & \texttt{IFQL} & \texttt{\color{myblue}FQL} &\texttt{\color{myred}Deflow} \\
\midrule

\texttt{antmaze-large-navigate-singletask-task1-v0 (*)} & $0$ {\tiny $\pm 0$} & $48$ {\tiny $\pm 9$} & $\mathbf{91}$ {\tiny $\pm 10$} & $0$ {\tiny $\pm 0$} & $0$ {\tiny $\pm 0$} & $42$ {\tiny $\pm 7$} & $1$ {\tiny $\pm 1$} & $70$ {\tiny $\pm 20$} & $24$ {\tiny $\pm 17$} & $80$ {\tiny $\pm 8$}& $\mathbf{88}$ {\tiny $\pm 6$} \\
\texttt{antmaze-large-navigate-singletask-task2-v0} & $6$ {\tiny $\pm 3$} & $42$ {\tiny $\pm 6$} & $\mathbf{88}$ {\tiny $\pm 4$} & $14$ {\tiny $\pm 8$} & $4$ {\tiny $\pm 4$} & $1$ {\tiny $\pm 1$} & $0$ {\tiny $\pm 1$} & $35$ {\tiny $\pm 12$} & $8$ {\tiny $\pm 3$} & $57$ {\tiny $\pm 10$} & $71$ {\tiny $\pm 10$}\\
\texttt{antmaze-large-navigate-singletask-task3-v0} & $29$ {\tiny $\pm 5$} & $72$ {\tiny $\pm 7$} & $51$ {\tiny $\pm 18$} & $26$ {\tiny $\pm 8$} & $3$ {\tiny $\pm 2$} & $49$ {\tiny $\pm 10$} & $12$ {\tiny $\pm 4$} & $83$ {\tiny $\pm 15$} & $52$ {\tiny $\pm 17$} & $\mathbf{93}$ {\tiny $\pm 3$} & $\mathbf{88}$ {\tiny $\pm 5$} \\
\texttt{antmaze-large-navigate-singletask-task4-v0} & $8$ {\tiny $\pm 3$} & $51$ {\tiny $\pm 9$} & $\mathbf{84}$ {\tiny $\pm 7$} & $62$ {\tiny $\pm 25$} & $45$ {\tiny $\pm 19$} & $17$ {\tiny $\pm 6$} & $10$ {\tiny $\pm 3$} & $37$ {\tiny $\pm 18$} & $18$ {\tiny $\pm 8$} & $\mathbf{80}$ {\tiny $\pm 4$} & $77$ {\tiny $\pm 5$} \\
\texttt{antmaze-large-navigate-singletask-task5-v0} & $10$ {\tiny $\pm 3$} & $54$ {\tiny $\pm 22$} & $\mathbf{90}$ {\tiny $\pm 2$} & $2$ {\tiny $\pm 2$} & $1$ {\tiny $\pm 1$} & $55$ {\tiny $\pm 6$} & $9$ {\tiny $\pm 5$} & $76$ {\tiny $\pm 8$} & $38$ {\tiny $\pm 18$} & $83$ {\tiny $\pm 4$} & $84$ {\tiny $\pm 3$}\\
\midrule
\texttt{antmaze-giant-navigate-singletask-task1-v0 (*)} & $0$ {\tiny $\pm 0$} & $0$ {\tiny $\pm 0$} & $\mathbf{27}$ {\tiny $\pm 22$} & $0$ {\tiny $\pm 0$} & $0$ {\tiny $\pm 0$} & $0$ {\tiny $\pm 0$} & $0$ {\tiny $\pm 0$} & $0$ {\tiny $\pm 1$} & $0$ {\tiny $\pm 0$} & $4$ {\tiny $\pm 5$} & $4$ {\tiny $\pm 4$}\\
\texttt{antmaze-giant-navigate-singletask-task2-v0} & $0$ {\tiny $\pm 0$} & $1$ {\tiny $\pm 1$} & $16$ {\tiny $\pm 17$} & $0$ {\tiny $\pm 0$} & $0$ {\tiny $\pm 0$} & $0$ {\tiny $\pm 0$} & $0$ {\tiny $\pm 0$} & $4$ {\tiny $\pm 7$} & $0$ {\tiny $\pm 0$} & $9$ {\tiny $\pm 7$} & $\mathbf{36}$ {\tiny $\pm 13$} \\
\texttt{antmaze-giant-navigate-singletask-task3-v0} & $0$ {\tiny $\pm 0$} & $0$ {\tiny $\pm 0$} & $\mathbf{34}$ {\tiny $\pm 22$} & $0$ {\tiny $\pm 0$} & $0$ {\tiny $\pm 0$} & $0$ {\tiny $\pm 0$} & $0$ {\tiny $\pm 0$} & $0$ {\tiny $\pm 0$} & $0$ {\tiny $\pm 0$} & $0$ {\tiny $\pm 1$}& $2$ {\tiny $\pm 3$} \\
\texttt{antmaze-giant-navigate-singletask-task4-v0} & $0$ {\tiny $\pm 0$} & $0$ {\tiny $\pm 0$} & $5$ {\tiny $\pm 12$} & $0$ {\tiny $\pm 0$} & $0$ {\tiny $\pm 0$} & $0$ {\tiny $\pm 0$} & $0$ {\tiny $\pm 0$} & $9$ {\tiny $\pm 4$} & $0$ {\tiny $\pm 0$} & $\mathbf{14}$ {\tiny $\pm 23$}& $5$ {\tiny $\pm 8$} \\
\texttt{antmaze-giant-navigate-singletask-task5-v0} & $1$ {\tiny $\pm 1$} & $19$ {\tiny $\pm 7$} & $\mathbf{49}$ {\tiny $\pm 22$} & $0$ {\tiny $\pm 1$} & $0$ {\tiny $\pm 0$} & $0$ {\tiny $\pm 0$} & $0$ {\tiny $\pm 0$} & $6$ {\tiny $\pm 10$} & $13$ {\tiny $\pm 9$} & $16$ {\tiny $\pm 28$}& $16$ {\tiny $\pm 19$} \\
\midrule
\texttt{humanoidmaze-medium-navigate-singletask-task1-v0 (*)} & $1$ {\tiny $\pm 0$} & $32$ {\tiny $\pm 7$} & $16$ {\tiny $\pm 9$} & $1$ {\tiny $\pm 1$} & $0$ {\tiny $\pm 0$} & $38$ {\tiny $\pm 19$} & $6$ {\tiny $\pm 2$} & $25$ {\tiny $\pm 8$} & $\mathbf{69}$ {\tiny $\pm 19$} & $19$ {\tiny $\pm 12$}& $21$ {\tiny $\pm 11$} \\
\texttt{humanoidmaze-medium-navigate-singletask-task2-v0} & $1$ {\tiny $\pm 0$} & $41$ {\tiny $\pm 9$} & $18$ {\tiny $\pm 16$} & $1$ {\tiny $\pm 1$} & $1$ {\tiny $\pm 1$} & $47$ {\tiny $\pm 35$} & $40$ {\tiny $\pm 2$} & $76$ {\tiny $\pm 10$} & $85$ {\tiny $\pm 11$} & $\mathbf{94}$ {\tiny $\pm 3$}& $69$ {\tiny $\pm 5$} \\
\texttt{humanoidmaze-medium-navigate-singletask-task3-v0} & $6$ {\tiny $\pm 2$} & $25$ {\tiny $\pm 5$} & $36$ {\tiny $\pm 13$} & $0$ {\tiny $\pm 1$} & $2$ {\tiny $\pm 1$} & $\mathbf{83}$ {\tiny $\pm 18$} & $19$ {\tiny $\pm 2$} & $27$ {\tiny $\pm 11$} & $49$ {\tiny $\pm 49$} & $74$ {\tiny $\pm 18$}& $63$ {\tiny $\pm 14$} \\
\texttt{humanoidmaze-medium-navigate-singletask-task4-v0} & $0$ {\tiny $\pm 0$} & $0$ {\tiny $\pm 1$} & $\mathbf{15}$ {\tiny $\pm 16$} & $1$ {\tiny $\pm 1$} & $1$ {\tiny $\pm 1$} & $5$ {\tiny $\pm 4$} & $1$ {\tiny $\pm 1$} & $1$ {\tiny $\pm 2$} & $1$ {\tiny $\pm 1$} & $3$ {\tiny $\pm 4$}& $6$ {\tiny $\pm 7$} \\
\texttt{humanoidmaze-medium-navigate-singletask-task5-v0} & $2$ {\tiny $\pm 1$} & $66$ {\tiny $\pm 4$} & $24$ {\tiny $\pm 20$} & $1$ {\tiny $\pm 1$} & $3$ {\tiny $\pm 3$} & $91$ {\tiny $\pm 5$} & $31$ {\tiny $\pm 7$} & $63$ {\tiny $\pm 9$} & $\mathbf{98}$ {\tiny $\pm 2$} & $\mathbf{97}$ {\tiny $\pm 2$}& $81$ {\tiny $\pm 7$} \\
\midrule
\texttt{humanoidmaze-large-navigate-singletask-task1-v0 (*)} & $0$ {\tiny $\pm 0$} & $3$ {\tiny $\pm 1$} & $2$ {\tiny $\pm 1$} & $0$ {\tiny $\pm 0$} & $0$ {\tiny $\pm 0$} & $1$ {\tiny $\pm 1$} & $0$ {\tiny $\pm 0$} & $0$ {\tiny $\pm 1$} & $6$ {\tiny $\pm 2$} & $\mathbf{7}$ {\tiny $\pm 6$}& $3$ {\tiny $\pm 4$} \\
\texttt{humanoidmaze-large-navigate-singletask-task2-v0} & $\mathbf{0}$ {\tiny $\pm 0$} & $\mathbf{0}$ {\tiny $\pm 0$} & $\mathbf{0}$ {\tiny $\pm 0$} & $\mathbf{0}$ {\tiny $\pm 0$} & $\mathbf{0}$ {\tiny $\pm 0$} & $\mathbf{0}$ {\tiny $\pm 0$} & $\mathbf{0}$ {\tiny $\pm 0$} & $\mathbf{0}$ {\tiny $\pm 0$} & $\mathbf{0}$ {\tiny $\pm 0$} & $\mathbf{0}$ {\tiny $\pm 0$}& $\mathbf{0}$ {\tiny $\pm 4$} \\
\texttt{humanoidmaze-large-navigate-singletask-task3-v0} & $1$ {\tiny $\pm 1$} & $7$ {\tiny $\pm 3$} & $8$ {\tiny $\pm 4$} & $3$ {\tiny $\pm 1$} & $1$ {\tiny $\pm 1$} & $2$ {\tiny $\pm 3$} & $1$ {\tiny $\pm 1$} & $10$ {\tiny $\pm 2$} & $\mathbf{48}$ {\tiny $\pm 10$} & $11$ {\tiny $\pm 7$}& $10$ {\tiny $\pm 6$} \\
\texttt{humanoidmaze-large-navigate-singletask-task4-v0} & $1$ {\tiny $\pm 0$} & $1$ {\tiny $\pm 0$} & $1$ {\tiny $\pm 1$} & $0$ {\tiny $\pm 0$} & $0$ {\tiny $\pm 0$} & $0$ {\tiny $\pm 1$} & $0$ {\tiny $\pm 0$} & $0$ {\tiny $\pm 0$} & $1$ {\tiny $\pm 1$} & $2$ {\tiny $\pm 3$}& $\mathbf{8}$ {\tiny $\pm 3$} \\
\texttt{humanoidmaze-large-navigate-singletask-task5-v0} & $0$ {\tiny $\pm 1$} & $1$ {\tiny $\pm 1$} & $\mathbf{2}$ {\tiny $\pm 2$} & $0$ {\tiny $\pm 0$} & $0$ {\tiny $\pm 0$} & $0$ {\tiny $\pm 0$} & $0$ {\tiny $\pm 0$} & $1$ {\tiny $\pm 1$} & $0$ {\tiny $\pm 0$} & $1$ {\tiny $\pm 3$}& $1$ {\tiny $\pm 1$} \\
\midrule
\texttt{antsoccer-arena-navigate-singletask-task1-v0} & $2$ {\tiny $\pm 1$} & $14$ {\tiny $\pm 5$} & $0$ {\tiny $\pm 0$} & $44$ {\tiny $\pm 12$} & $2$ {\tiny $\pm 1$} & $1$ {\tiny $\pm 3$} & $22$ {\tiny $\pm 2$} & $17$ {\tiny $\pm 3$} & $61$ {\tiny $\pm 25$} & $77$ {\tiny $\pm 4$}& $\mathbf{84}$ {\tiny $\pm 7$} \\
\texttt{antsoccer-arena-navigate-singletask-task2-v0} & $2$ {\tiny $\pm 2$} & $17$ {\tiny $\pm 7$} & $0$ {\tiny $\pm 1$} & $15$ {\tiny $\pm 12$} & $3$ {\tiny $\pm 1$} & $0$ {\tiny $\pm 0$} & $8$ {\tiny $\pm 1$} & $8$ {\tiny $\pm 2$} & $75$ {\tiny $\pm 3$} & $\mathbf{88}$ {\tiny $\pm 3$}& $\mathbf{89}$ {\tiny $\pm 4$} \\
\texttt{antsoccer-arena-navigate-singletask-task3-v0} & $0$ {\tiny $\pm 0$} & $6$ {\tiny $\pm 4$} & $0$ {\tiny $\pm 0$} & $0$ {\tiny $\pm 0$} & $0$ {\tiny $\pm 0$} & $8$ {\tiny $\pm 19$} & $11$ {\tiny $\pm 5$} & $16$ {\tiny $\pm 3$} & $14$ {\tiny $\pm 22$} & $\mathbf{61}$ {\tiny $\pm 6$}& $56$ {\tiny $\pm 7$} \\
\texttt{antsoccer-arena-navigate-singletask-task4-v0 (*)} & $1$ {\tiny $\pm 0$} & $3$ {\tiny $\pm 2$} & $0$ {\tiny $\pm 0$} & $0$ {\tiny $\pm 1$} & $0$ {\tiny $\pm 0$} & $0$ {\tiny $\pm 0$} & $12$ {\tiny $\pm 3$} & $24$ {\tiny $\pm 4$} & $16$ {\tiny $\pm 9$} & $39$ {\tiny $\pm 6$}& $\mathbf{43}$ {\tiny $\pm 5$} \\
\texttt{antsoccer-arena-navigate-singletask-task5-v0} & $0$ {\tiny $\pm 0$} & $2$ {\tiny $\pm 2$} & $0$ {\tiny $\pm 0$} & $0$ {\tiny $\pm 0$} & $0$ {\tiny $\pm 0$} & $0$ {\tiny $\pm 0$} & $9$ {\tiny $\pm 2$} & $15$ {\tiny $\pm 4$} & $0$ {\tiny $\pm 1$} & $36$ {\tiny $\pm 9$}& $\mathbf{39}$ {\tiny $\pm 9$} \\
\midrule
\texttt{cube-single-play-singletask-task1-v0} & $10$ {\tiny $\pm 5$} & $88$ {\tiny $\pm 3$} & $89$ {\tiny $\pm 5$} & $\mathbf{95}$ {\tiny $\pm 2$} & $89$ {\tiny $\pm 7$} & $77$ {\tiny $\pm 28$} & $81$ {\tiny $\pm 9$} & $73$ {\tiny $\pm 33$} & $79$ {\tiny $\pm 4$} & $\mathbf{97}$ {\tiny $\pm 2$}& $\mathbf{96}$ {\tiny $\pm 3$} \\
\texttt{cube-single-play-singletask-task2-v0 (*)} & $3$ {\tiny $\pm 1$} & $85$ {\tiny $\pm 8$} & $92$ {\tiny $\pm 4$} & $\mathbf{96}$ {\tiny $\pm 2$} & $82$ {\tiny $\pm 16$} & $80$ {\tiny $\pm 30$} & $81$ {\tiny $\pm 9$} & $83$ {\tiny $\pm 13$} & $73$ {\tiny $\pm 3$} & $\mathbf{97}$ {\tiny $\pm 2$}& $\mathbf{99}$ {\tiny $\pm 1$} \\
\texttt{cube-single-play-singletask-task3-v0} & $9$ {\tiny $\pm 3$} & $91$ {\tiny $\pm 5$} & $93$ {\tiny $\pm 3$} & $\mathbf{99}$ {\tiny $\pm 1$} & $\mathbf{96}$ {\tiny $\pm 2$} & $\mathbf{98}$ {\tiny $\pm 1$} & $87$ {\tiny $\pm 4$} & $82$ {\tiny $\pm 12$} & $88$ {\tiny $\pm 4$} & $\mathbf{98}$ {\tiny $\pm 2$}& $\mathbf{100}$ {\tiny $\pm 1$} \\
\texttt{cube-single-play-singletask-task4-v0} & $2$ {\tiny $\pm 1$} & $73$ {\tiny $\pm 6$} & $\mathbf{92}$ {\tiny $\pm 3$} & $\mathbf{93}$ {\tiny $\pm 4$} & $70$ {\tiny $\pm 18$} & $\mathbf{91}$ {\tiny $\pm 2$} & $79$ {\tiny $\pm 6$} & $79$ {\tiny $\pm 20$} & $79$ {\tiny $\pm 6$} & $\mathbf{94}$ {\tiny $\pm 3$}& $\mathbf{95}$ {\tiny $\pm 4$} \\
\texttt{cube-single-play-singletask-task5-v0} & $3$ {\tiny $\pm 3$} & $78$ {\tiny $\pm 9$} & $87$ {\tiny $\pm 8$} & $\mathbf{90}$ {\tiny $\pm 6$} & $61$ {\tiny $\pm 12$} & $80$ {\tiny $\pm 20$} & $78$ {\tiny $\pm 10$} & $76$ {\tiny $\pm 33$} & $77$ {\tiny $\pm 7$} & $\mathbf{93}$ {\tiny $\pm 3$}& $\mathbf{92}$ {\tiny $\pm 5$} \\
\bottomrule
\end{tabular}
\end{threeparttable}
}
\end{table}
\clearpage
\begin{table}[H]
\ContinuedFloat
\caption{\textbf{Full offline RL results (continued).}}
\centering
\setlength{\tabcolsep}{2.5pt}
\resizebox{\textwidth}{!}{%
\begin{threeparttable}
\begin{tabular}{lccccccccccc}
\toprule
\multicolumn{1}{c}{} & \multicolumn{3}{c}{\texttt{Gaussian Policies}} & \multicolumn{3}{c}{\texttt{Diffusion Policies}} & \multicolumn{5}{c}{\texttt{Flow Policies}} \\
\cmidrule(lr){2-4} \cmidrule(lr){5-7} \cmidrule(lr){8-12}
\texttt{Task} & \texttt{BC} & \texttt{IQL} & \texttt{ReBRAC} & \texttt{IDQL} & \texttt{SRPO} & \texttt{CAC} & \texttt{FAWAC} & \texttt{FBRAC} & \texttt{IFQL} & \texttt{\color{myblue}FQL} &\texttt{\color{myred}Deflow} \\
\midrule

\texttt{cube-double-play-singletask-task1-v0} & $8$ {\tiny $\pm 3$} & $27$ {\tiny $\pm 5$} & $45$ {\tiny $\pm 6$} & $39$ {\tiny $\pm 19$} & $7$ {\tiny $\pm 6$} & $21$ {\tiny $\pm 8$} & $21$ {\tiny $\pm 7$} & $47$ {\tiny $\pm 11$} & $35$ {\tiny $\pm 9$} & $\mathbf{61}$ {\tiny $\pm 9$}& $\mathbf{65}$ {\tiny $\pm 6$} \\
\texttt{cube-double-play-singletask-task2-v0 (*)} & $0$ {\tiny $\pm 0$} & $1$ {\tiny $\pm 1$} & $7$ {\tiny $\pm 3$} & $16$ {\tiny $\pm 10$} & $0$ {\tiny $\pm 0$} & $2$ {\tiny $\pm 2$} & $2$ {\tiny $\pm 1$} & $22$ {\tiny $\pm 12$} & $9$ {\tiny $\pm 5$} & $36$ {\tiny $\pm 6$}& $\mathbf{49}$ {\tiny $\pm 13$} \\
\texttt{cube-double-play-singletask-task3-v0} & $0$ {\tiny $\pm 0$} & $0$ {\tiny $\pm 0$} & $4$ {\tiny $\pm 1$} & $17$ {\tiny $\pm 8$} & $0$ {\tiny $\pm 1$} & $3$ {\tiny $\pm 1$} & $1$ {\tiny $\pm 1$} & $4$ {\tiny $\pm 2$} & $8$ {\tiny $\pm 5$} & $22$ {\tiny $\pm 5$}& $\mathbf{34}$ {\tiny $\pm 15$} \\
\texttt{cube-double-play-singletask-task4-v0} & $0$ {\tiny $\pm 0$} & $0$ {\tiny $\pm 0$} & $1$ {\tiny $\pm 1$} & $0$ {\tiny $\pm 1$} & $0$ {\tiny $\pm 0$} & $0$ {\tiny $\pm 1$} & $0$ {\tiny $\pm 0$} & $0$ {\tiny $\pm 1$} & $1$ {\tiny $\pm 1$} & $5$ {\tiny $\pm 2$}& $\mathbf{7}$ {\tiny $\pm 4$} \\
\texttt{cube-double-play-singletask-task5-v0} & $0$ {\tiny $\pm 0$} & $4$ {\tiny $\pm 3$} & $4$ {\tiny $\pm 2$} & $1$ {\tiny $\pm 1$} & $0$ {\tiny $\pm 0$} & $3$ {\tiny $\pm 2$} & $2$ {\tiny $\pm 1$} & $2$ {\tiny $\pm 2$} & $17$ {\tiny $\pm 6$} & $19$ {\tiny $\pm 10$}& $\mathbf{46}$ {\tiny $\pm 11$} \\
\midrule
\texttt{scene-play-singletask-task1-v0} & $19$ {\tiny $\pm 6$} & $94$ {\tiny $\pm 3$} & $\mathbf{95}$ {\tiny $\pm 2$} & $\mathbf{100}$ {\tiny $\pm 0$} & $94$ {\tiny $\pm 4$} & $\mathbf{100}$ {\tiny $\pm 1$} & $87$ {\tiny $\pm 8$} & $\mathbf{96}$ {\tiny $\pm 8$} & $\mathbf{98}$ {\tiny $\pm 3$} & $\mathbf{100}$ {\tiny $\pm 0$}& $\mathbf{98}$ {\tiny $\pm 2$} \\
\texttt{scene-play-singletask-task2-v0 (*)} & $1$ {\tiny $\pm 1$} & $12$ {\tiny $\pm 3$} & $50$ {\tiny $\pm 13$} & $33$ {\tiny $\pm 14$} & $2$ {\tiny $\pm 2$} & $50$ {\tiny $\pm 40$} & $18$ {\tiny $\pm 8$} & $46$ {\tiny $\pm 10$} & $0$ {\tiny $\pm 0$} & $\mathbf{76}$ {\tiny $\pm 9$}& $59$ {\tiny $\pm 17$} \\
\texttt{scene-play-singletask-task3-v0} & $1$ {\tiny $\pm 1$} & $32$ {\tiny $\pm 7$} & $55$ {\tiny $\pm 16$} & $\mathbf{94}$ {\tiny $\pm 4$} & $4$ {\tiny $\pm 4$} & $49$ {\tiny $\pm 16$} & $38$ {\tiny $\pm 9$} & $78$ {\tiny $\pm 14$} & $54$ {\tiny $\pm 19$} & $\mathbf{98}$ {\tiny $\pm 1$}& $85$ {\tiny $\pm 5$} \\
\texttt{scene-play-singletask-task4-v0} & $2$ {\tiny $\pm 2$} & $0$ {\tiny $\pm 1$} & $3$ {\tiny $\pm 3$} & $4$ {\tiny $\pm 3$} & $0$ {\tiny $\pm 0$} & $0$ {\tiny $\pm 0$} & $6$ {\tiny $\pm 1$} & $4$ {\tiny $\pm 4$} & $0$ {\tiny $\pm 0$} & $5$ {\tiny $\pm 1$}& $\mathbf{13}$ {\tiny $\pm 8$} \\
\texttt{scene-play-singletask-task5-v0} & $\mathbf{0}$ {\tiny $\pm 0$} & $\mathbf{0}$ {\tiny $\pm 0$} & $\mathbf{0}$ {\tiny $\pm 0$} & $\mathbf{0}$ {\tiny $\pm 0$} & $\mathbf{0}$ {\tiny $\pm 0$} & $\mathbf{0}$ {\tiny $\pm 0$} & $\mathbf{0}$ {\tiny $\pm 0$} & $\mathbf{0}$ {\tiny $\pm 0$} & $\mathbf{0}$ {\tiny $\pm 0$} & $\mathbf{0}$ {\tiny $\pm 0$}& $\mathbf{0}$ {\tiny $\pm 0$} \\
\midrule
\texttt{puzzle-3x3-play-singletask-task1-v0} & $5$ {\tiny $\pm 2$} & $33$ {\tiny $\pm 6$} & $\mathbf{97}$ {\tiny $\pm 4$} & $52$ {\tiny $\pm 12$} & $89$ {\tiny $\pm 5$} & $\mathbf{97}$ {\tiny $\pm 2$} & $25$ {\tiny $\pm 9$} & $63$ {\tiny $\pm 19$} & $\mathbf{94}$ {\tiny $\pm 3$} & $90$ {\tiny $\pm 4$}& $89$ {\tiny $\pm 7$} \\
\texttt{puzzle-3x3-play-singletask-task2-v0} & $1$ {\tiny $\pm 1$} & $4$ {\tiny $\pm 3$} & $1$ {\tiny $\pm 1$} & $0$ {\tiny $\pm 1$} & $0$ {\tiny $\pm 1$} & $0$ {\tiny $\pm 0$} & $4$ {\tiny $\pm 2$} & $2$ {\tiny $\pm 2$} & $1$ {\tiny $\pm 2$} & $16$ {\tiny $\pm 5$}& $\mathbf{38}$ {\tiny $\pm 11$} \\
\texttt{puzzle-3x3-play-singletask-task3-v0} & $1$ {\tiny $\pm 1$} & $3$ {\tiny $\pm 2$} & $3$ {\tiny $\pm 1$} & $0$ {\tiny $\pm 0$} & $0$ {\tiny $\pm 0$} & $0$ {\tiny $\pm 0$} & $1$ {\tiny $\pm 0$} & $1$ {\tiny $\pm 1$} & $0$ {\tiny $\pm 0$} & $10$ {\tiny $\pm 3$}& $\mathbf{35}$ {\tiny $\pm 11$} \\
\texttt{puzzle-3x3-play-singletask-task4-v0 (*)} & $1$ {\tiny $\pm 1$} & $2$ {\tiny $\pm 1$} & $2$ {\tiny $\pm 1$} & $0$ {\tiny $\pm 0$} & $0$ {\tiny $\pm 0$} & $0$ {\tiny $\pm 0$} & $1$ {\tiny $\pm 1$} & $2$ {\tiny $\pm 2$} & $0$ {\tiny $\pm 0$} & $16$ {\tiny $\pm 5$}& $\mathbf{23}$ {\tiny $\pm 8$} \\
\texttt{puzzle-3x3-play-singletask-task5-v0} & $1$ {\tiny $\pm 0$} & $3$ {\tiny $\pm 2$} & $5$ {\tiny $\pm 3$} & $0$ {\tiny $\pm 0$} & $0$ {\tiny $\pm 0$} & $0$ {\tiny $\pm 0$} & $1$ {\tiny $\pm 1$} & $2$ {\tiny $\pm 2$} & $0$ {\tiny $\pm 0$} & $16$ {\tiny $\pm 3$}& $31$ {\tiny $\pm 8$} \\
\midrule
\texttt{puzzle-4x4-play-singletask-task1-v0} & $1$ {\tiny $\pm 1$} & $12$ {\tiny $\pm 2$} & $26$ {\tiny $\pm 4$} & $\mathbf{48}$ {\tiny $\pm 5$} & $24$ {\tiny $\pm 9$} & $44$ {\tiny $\pm 10$} & $1$ {\tiny $\pm 2$} & $32$ {\tiny $\pm 9$} & $\mathbf{49}$ {\tiny $\pm 9$} & $34$ {\tiny $\pm 8$}& $23$ {\tiny $\pm 3$} \\
\texttt{puzzle-4x4-play-singletask-task2-v0} & $0$ {\tiny $\pm 0$} & $7$ {\tiny $\pm 4$} & $12$ {\tiny $\pm 4$} & $14$ {\tiny $\pm 5$} & $0$ {\tiny $\pm 1$} & $0$ {\tiny $\pm 0$} & $0$ {\tiny $\pm 1$} & $5$ {\tiny $\pm 3$} & $4$ {\tiny $\pm 4$} & $\mathbf{16}$ {\tiny $\pm 5$}& $9$ {\tiny $\pm 2$} \\
\texttt{puzzle-4x4-play-singletask-task3-v0} & $0$ {\tiny $\pm 0$} & $9$ {\tiny $\pm 3$} & $15$ {\tiny $\pm 3$} & $34$ {\tiny $\pm 5$} & $21$ {\tiny $\pm 10$} & $29$ {\tiny $\pm 12$} & $1$ {\tiny $\pm 1$} & $20$ {\tiny $\pm 10$} & $\mathbf{50}$ {\tiny $\pm 14$} & $18$ {\tiny $\pm 5$}& $9$ {\tiny $\pm 4$} \\
\texttt{puzzle-4x4-play-singletask-task4-v0 (*)} & $0$ {\tiny $\pm 0$} & $5$ {\tiny $\pm 2$} & $10$ {\tiny $\pm 3$} & $\mathbf{26}$ {\tiny $\pm 6$} & $7$ {\tiny $\pm 4$} & $1$ {\tiny $\pm 1$} & $0$ {\tiny $\pm 0$} & $5$ {\tiny $\pm 1$} & $21$ {\tiny $\pm 11$} & $11$ {\tiny $\pm 3$}& $7$ {\tiny $\pm 4$} \\
\texttt{puzzle-4x4-play-singletask-task5-v0} & $0$ {\tiny $\pm 0$} & $4$ {\tiny $\pm 1$} & $7$ {\tiny $\pm 3$} & $\mathbf{24}$ {\tiny $\pm 11$} & $1$ {\tiny $\pm 1$} & $0$ {\tiny $\pm 0$} & $0$ {\tiny $\pm 1$} & $4$ {\tiny $\pm 3$} & $2$ {\tiny $\pm 2$} & $7$ {\tiny $\pm 3$}& $7$ {\tiny $\pm 4$} \\
\midrule
\texttt{antmaze-umaze-v2} & $55$ & $77$ & $\mathbf{98}$ & $\mathbf{94}$ & $\mathbf{97}$ & $66$ {\tiny $\pm 5$} & $90$ {\tiny $\pm 6$} & $\mathbf{94}$ {\tiny $\pm 3$} & $92$ {\tiny $\pm 6$} & $\mathbf{96}$ {\tiny $\pm 2$}& $\mathbf{96}$ {\tiny $\pm 3$} \\
\texttt{antmaze-umaze-diverse-v2} & $47$ & $54$ & $84$ & $80$ & $82$ & $66$ {\tiny $\pm 11$} & $55$ {\tiny $\pm 7$} & $82$ {\tiny $\pm 9$} & $62$ {\tiny $\pm 12$} & $\mathbf{89}$ {\tiny $\pm 5$}& $\mathbf{87}$ {\tiny $\pm 8$} \\
\texttt{antmaze-medium-play-v2} & $0$ & $66$ & $\mathbf{90}$ & $84$ & $81$ & $49$ {\tiny $\pm 24$} & $52$ {\tiny $\pm 12$} & $77$ {\tiny $\pm 7$} & $56$ {\tiny $\pm 15$} & $78$ {\tiny $\pm 7$}& $80$ {\tiny $\pm 7$} \\
\texttt{antmaze-medium-diverse-v2} & $1$ & $74$ & $\mathbf{84}$ & $\mathbf{85}$ & $75$ & $0$ {\tiny $\pm 1$} & $44$ {\tiny $\pm 15$} & $77$ {\tiny $\pm 6$} & $60$ {\tiny $\pm 25$} & $71$ {\tiny $\pm 13$}& $76$ {\tiny $\pm 19$} \\
\texttt{antmaze-large-play-v2} & $0$ & $42$ & $52$ & $64$ & $54$ & $0$ {\tiny $\pm 0$} & $10$ {\tiny $\pm 6$} & $32$ {\tiny $\pm 21$} & $55$ {\tiny $\pm 9$} & $\mathbf{84}$ {\tiny $\pm 7$}& $\mathbf{78}$ {\tiny $\pm 6$} \\
\texttt{antmaze-large-diverse-v2} & $0$ & $30$ & $64$ & $68$ & $54$ & $0$ {\tiny $\pm 0$} & $16$ {\tiny $\pm 10$} & $20$ {\tiny $\pm 17$} & $64$ {\tiny $\pm 8$} & $\mathbf{83}$ {\tiny $\pm 4$}& $78$ {\tiny $\pm 6$} \\
\midrule
\texttt{pen-human-v1} & $71$ & $78$ & $\mathbf{103}$ & $76$ {\tiny $\pm 10$} & $69$ {\tiny $\pm 7$} & $64$ {\tiny $\pm 8$} & $67$ {\tiny $\pm 5$} & $77$ {\tiny $\pm 7$} & $71$ {\tiny $\pm 12$} & $53$ {\tiny $\pm 6$}& $61$ {\tiny $\pm 9$} \\
\texttt{pen-cloned-v1} & $52$ & $83$ & $\mathbf{103}$ & $64$ {\tiny $\pm 7$} & $61$ {\tiny $\pm 7$} & $56$ {\tiny $\pm 10$} & $62$ {\tiny $\pm 10$} & $67$ {\tiny $\pm 9$} & $80$ {\tiny $\pm 11$} & $74$ {\tiny $\pm 11$}& $75$ {\tiny $\pm 4$} \\
\texttt{pen-expert-v1} & $110$ & $128$ & $\mathbf{152}$ & $140$ {\tiny $\pm 6$} & $134$ {\tiny $\pm 4$} & $103$ {\tiny $\pm 9$} & $118$ {\tiny $\pm 6$} & $119$ {\tiny $\pm 7$} & $139$ {\tiny $\pm 5$} & $142$ {\tiny $\pm 6$}& $138$ {\tiny $\pm 5$} \\
\texttt{door-human-v1} & $2$ & $3$ & $-0$ & $6$ {\tiny $\pm 2$} & $3$ {\tiny $\pm 3$} & $5$ {\tiny $\pm 2$} & $2$ {\tiny $\pm 1$} & $4$ {\tiny $\pm 2$} & $\mathbf{7}$ {\tiny $\pm 2$} & $0$ {\tiny $\pm 0$}& $1$ {\tiny $\pm 0$} \\
\texttt{door-cloned-v1} & $-0$ & $\mathbf{3}$ & $0$ & $0$ {\tiny $\pm 0$} & $0$ {\tiny $\pm 0$} & $1$ {\tiny $\pm 0$} & $0$ {\tiny $\pm 1$} & $0$ {\tiny $\pm 0$} & $2$ {\tiny $\pm 2$} & $2$ {\tiny $\pm 1$}& $2$ {\tiny $\pm 1$} \\
\texttt{door-expert-v1} & $\mathbf{105}$ & $\mathbf{107}$ & $\mathbf{106}$ & $\mathbf{105}$ {\tiny $\pm 1$} & $\mathbf{105}$ {\tiny $\pm 0$} & $98$ {\tiny $\pm 3$} & $\mathbf{103}$ {\tiny $\pm 1$} & $\mathbf{104}$ {\tiny $\pm 1$} & $\mathbf{104}$ {\tiny $\pm 2$} & $\mathbf{104}$ {\tiny $\pm 1$}& $\mathbf{104}$ {\tiny $\pm 1$} \\
\texttt{hammer-human-v1} & $\mathbf{3}$ & $2$ & $0$ & $2$ {\tiny $\pm 1$} & $1$ {\tiny $\pm 1$} & $2$ {\tiny $\pm 0$} & $2$ {\tiny $\pm 1$} & $2$ {\tiny $\pm 1$} & $\mathbf{3}$ {\tiny $\pm 1$} & $1$ {\tiny $\pm 1$}& $1$ {\tiny $\pm 0$} \\
\texttt{hammer-cloned-v1} & $1$ & $2$ & $5$ & $2$ {\tiny $\pm 1$} & $2$ {\tiny $\pm 1$} & $1$ {\tiny $\pm 1$} & $1$ {\tiny $\pm 0$} & $2$ {\tiny $\pm 1$} & $2$ {\tiny $\pm 1$} & $\mathbf{11}$ {\tiny $\pm 9$}& $7$ {\tiny $\pm 4$} \\
\texttt{hammer-expert-v1} & $127$ & $\mathbf{129}$ & $\mathbf{134}$ & $125$ {\tiny $\pm 4$} & $127$ {\tiny $\pm 0$} & $92$ {\tiny $\pm 11$} & $118$ {\tiny $\pm 3$} & $119$ {\tiny $\pm 9$} & $117$ {\tiny $\pm 9$} & $125$ {\tiny $\pm 3$}& $127$ {\tiny $\pm 1$} \\
\texttt{relocate-human-v1} & $\mathbf{0}$ & $\mathbf{0}$ & $\mathbf{0}$ & $\mathbf{0}$ {\tiny $\pm 0$} & $\mathbf{0}$ {\tiny $\pm 0$} & $\mathbf{0}$ {\tiny $\pm 0$} & $\mathbf{0}$ {\tiny $\pm 0$} & $\mathbf{0}$ {\tiny $\pm 0$} & $\mathbf{0}$ {\tiny $\pm 0$} & $\mathbf{0}$ {\tiny $\pm 0$}& $\mathbf{0}$ {\tiny $\pm 0$} \\
\texttt{relocate-cloned-v1} & $-0$ & $0$ & $\mathbf{2}$ & $-0$ {\tiny $\pm 0$} & $-0$ {\tiny $\pm 0$} & $-0$ {\tiny $\pm 0$} & $-0$ {\tiny $\pm 0$} & $1$ {\tiny $\pm 1$} & $-0$ {\tiny $\pm 0$} & $-0$ {\tiny $\pm 0$}& $1$ {\tiny $\pm 8$} \\
\texttt{relocate-expert-v1} & $\mathbf{108}$ & $\mathbf{106}$ & $\mathbf{108}$ & $\mathbf{107}$ {\tiny $\pm 1$} & $\mathbf{106}$ {\tiny $\pm 2$} & $93$ {\tiny $\pm 6$} & $\mathbf{105}$ {\tiny $\pm 3$} & $\mathbf{105}$ {\tiny $\pm 2$} & $\mathbf{104}$ {\tiny $\pm 3$} & $\mathbf{107}$ {\tiny $\pm 1$}& $\mathbf{109}$ {\tiny $\pm 2$} \\
\midrule
\texttt{visual-cube-single-play-singletask-task1-v0}\tnote{1} & - & $70$ {\tiny $\pm 12$} & $\mathbf{83}$ {\tiny $\pm 6$} & - & - & - & - & $55$ {\tiny $\pm 8$} & $49$ {\tiny $\pm 7$} & $81$ {\tiny $\pm 12$}& $\mathbf{86}$ {\tiny $\pm 9$} \\
\texttt{visual-cube-double-play-singletask-task1-v0}\tnote{1} & - & $\mathbf{34}$ {\tiny $\pm 23$} & $4$ {\tiny $\pm 4$} & - & - & - & - & $6$ {\tiny $\pm 2$} & $8$ {\tiny $\pm 6$} & $21$ {\tiny $\pm 11$}& $\mathbf{53}$ {\tiny $\pm 7$} \\
\texttt{visual-scene-play-singletask-task1-v0}\tnote{1} & - & $\mathbf{97}$ {\tiny $\pm 2$} & $\mathbf{98}$ {\tiny $\pm 4$} & - & - & - & - & $46$ {\tiny $\pm 4$} & $86$ {\tiny $\pm 10$} & $\mathbf{98}$ {\tiny $\pm 3$}& $\mathbf{97}$ {\tiny $\pm 2$} \\
\texttt{visual-puzzle-3x3-play-singletask-task1-v0}\tnote{1} & - & $7$ {\tiny $\pm 15$} & $88$ {\tiny $\pm 4$} & - & - & - & - & $7$ {\tiny $\pm 2$} & $\mathbf{100}$ {\tiny $\pm 0$} & $94$ {\tiny $\pm 1$}& $\mathbf{97}$ {\tiny $\pm 2$} \\
\texttt{visual-puzzle-4x4-play-singletask-task1-v0}\tnote{1} & - & $0$ {\tiny $\pm 0$} & $26$ {\tiny $\pm 6$} & - & - & - & - & $0$ {\tiny $\pm 0$} & $8$ {\tiny $\pm 15$} & $\mathbf{33}$ {\tiny $\pm 6$}& $14$ {\tiny $\pm 2$} \\

\bottomrule
\end{tabular}
\begin{tablenotes}
\item[1] Due to the high computational cost of pixel-based tasks,
we selectively benchmark $5$ methods that achieve strong performance on state-based OGBench tasks.
\end{tablenotes}
\end{threeparttable}
}
\end{table}

\clearpage

\end{document}